\documentclass[lettersize,journal]{IEEEtran}
\usepackage{amsmath,amsfonts}
\usepackage{algorithmic}
\usepackage{algorithm}
\usepackage{array}
\usepackage[caption=false,font=normalsize,labelfont=sf,textfont=sf]{subfig}
\usepackage{textcomp}
\usepackage{stfloats}
\usepackage{url}
\usepackage{verbatim}
\usepackage{graphicx}
\usepackage{cite}

\usepackage{microtype}
\usepackage{booktabs}
\usepackage[capitalize,noabbrev]{cleveref}

\usepackage{multirow}
\usepackage{multicol}
\usepackage{pifont}
\usepackage{makecell}
\usepackage{bm}

\begin{document}

\title{NOAH: Learning Pairwise Object Category Attentions for Image Classification}

\author{Chao Li\textsuperscript{$\ast$},
        Aojun Zhou,
        and Anbang Yao$^\dagger$
\IEEEcompsocitemizethanks{
\IEEEcompsocthanksitem Chao Li and Anbang Yao are with Intel Labs China. E-mail: \{chao3.li, anbang.yao\}@intel.com
\IEEEcompsocthanksitem Aojun Zhou is with CUHK-SenseTime Joint Lab, The Chinese University of Hong Kong. Email: aojun.zhou@gmail.com
\IEEEcompsocthanksitem \textsuperscript{$\ast$}This work was done when Chao Li was an intern at Intel Labs China. Anbang Yao led the project and the paper writing. $^\dagger$ Corresponding author.%
}
}
\markboth{Journal of \LaTeX\ Class Files,~Vol.~14, No.~8, August~2021}%
{Shell \MakeLowercase{\textit{et al.}}: A Sample Article Using IEEEtran.cls for IEEE Journals}


\maketitle

\begin{abstract}
A modern deep neural network (DNN) for image classification tasks typically consists of two parts: a backbone for feature extraction, and a head for feature encoding and class predication. We observe that the head structures of mainstream DNNs adopt a similar feature encoding pipeline, exploiting global feature dependencies while disregarding local ones. In this paper, we revisit the feature encoding problem, and propose Non-glObal Attentive Head (NOAH) that relies on a new form of dot-product attention called pairwise object category attention (POCA), efficiently exploiting spatially dense category-specific attentions to augment classification performance. NOAH introduces 
a neat combination of feature split, transform and merge operations to learn POCAs at local to global scales. As a drop-in design, NOAH can be easily used to replace existing heads of various types of DNNs, improving classification performance while maintaining similar model efficiency. We validate the effectiveness of NOAH on ImageNet classification benchmark with 25 DNN architectures spanning convolutional neural networks, vision transformers and multi-layer perceptrons. In general, NOAH is able to significantly improve the performance of lightweight DNNs, e.g., showing $3.14\%$$|$$5.3\%$$|$$1.9\%$ top-1 accuracy improvement to MobileNetV2 ($0.5\times$)$|$Deit-Tiny ($0.5\times$)$|$gMLP-Tiny ($0.5\times$). NOAH also generalizes well when applied to medium-size and large-size DNNs. We further show that NOAH retains its efficacy on other popular multi-class and multi-label image classification benchmarks as well as in different training regimes, e.g., showing $3.6\%$$|$$1.1\%$ mAP improvement to large ResNet101$|$ViT-Large on MS-COCO dataset. Project page: https://github.com/OSVAI/NOAH.
\end{abstract}


\section{Introduction}
Image classification, a fundamental research topic in computer vision, has been actively studied for decades. 
Typically, it is formulated as either a multi-class or a multi-label classification problem. In multi-class classification scenarios such as face recognition and human action recognition, a model needs to predict a single label among a predefined set of classes for an input image. Comparatively, in multi-label classification scenarios such as vehicle attribute recognition and document classification, a model requires that an input image can be assigned with multiple labels. In the pre-deep-learning era, bag-of-feature classification models~\cite{sift,hog,spatial_pyramid_matching} were widely used. With the tremendous advances in deep learning, deep neural networks (DNNs) have become the predominant learning models for image classification, leading to a paradigm shift from hand-crafted feature design to neural architecture design.~\textit{In this paper, we focus on improving the performance of DNNs for both multi-class and multi-label image classification scenarios}.

\begin{figure*}[ht]
\vskip -0.in
	\begin{center}
		\includegraphics[width=.95\linewidth]{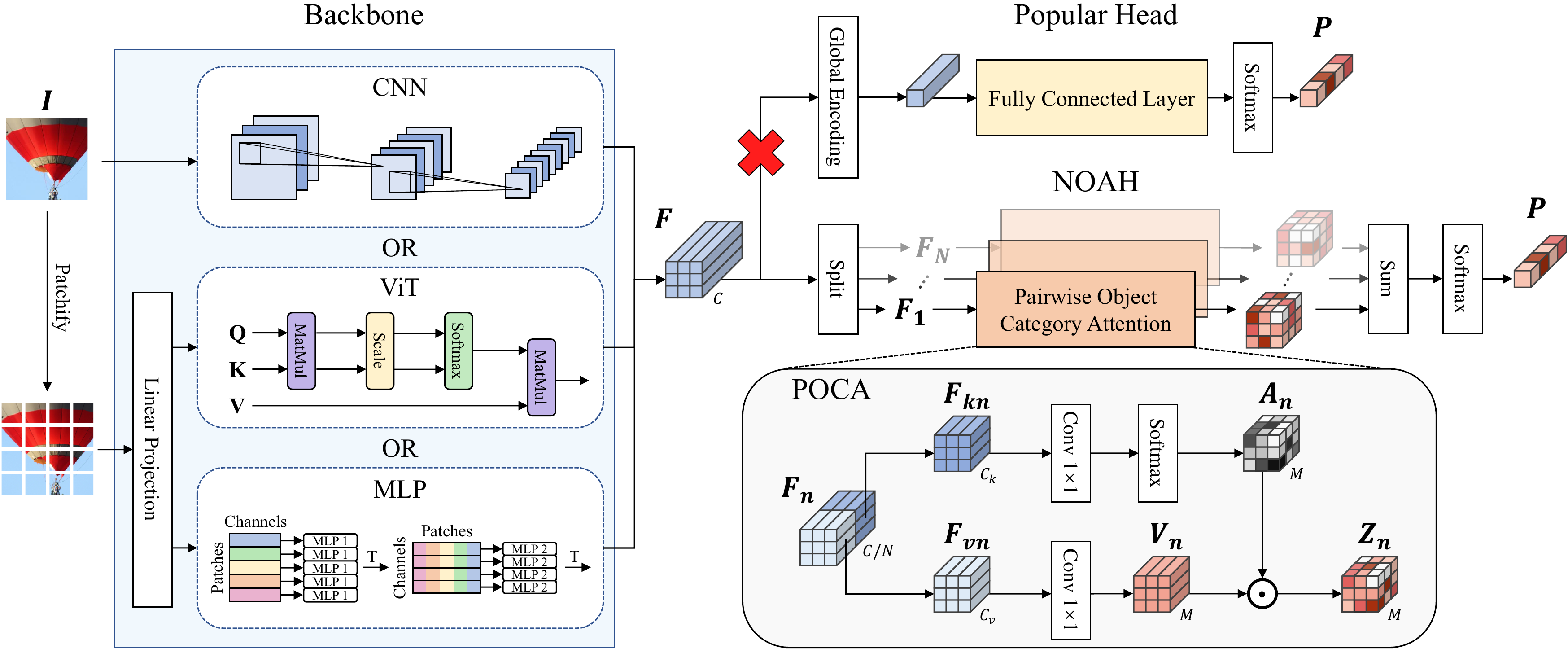}
	\end{center}
\vskip -0.in
	\caption{An architectural overview of DNN backbones appended with a Non-glObal Attentive Head (NOAH). Unlike the Popular Head based on global feature encoding, our NOAH relies on pairwise object category attentions (POCAs) learnt at local to global scales via a neat combination of feature split (two levels), transform and merge operations, taking the feature maps from the last layer of a backbone as the input.} 
	\label{fig:noah}
\vskip -0.in
\end{figure*}

Modern DNN architectures for image classification are constructed with a de-facto engineering pipeline that decomposes the network body into two parts: a backbone for feature extraction, and a head for feature encoding and class predication. Along with substantial research efforts in the backbone engineering, current classification DNN architectures have evolved into three major categories, namely convolutional neural networks (CNNs), vision transformers (ViTs) and multi-layer perceptrons (MLPs), primarily built on convolutional, self-attention and linear layers, respectively. Over the past decade, CNNs have been the go-to image classification models, which are known to have the inductive bias possessing the locality and translation equivariance across all convolutional layers. Early pioneering CNN models, AlexNet~\cite{alexnet} and VGG~\cite{vgg}, use a parameter-intensive head consisting of a max pooling layer, three fully connected (FC) layers and a softmax classifier. GoogLeNet~\cite{googlenet} replaces the max pooling layer by a global average pooling (GAP) layer based on NIN~\cite{nin}, and further shows that removing the first two FC layers does not incur accuracy drop yet enjoys significantly reduced model size. Consequently, subsequent CNNs~\cite{resnet,nas,regnet,convnext} mostly follow the head design of GoogLeNet. In 2021, Dosovitskiy et al.~\cite{vit} directly applied a pure transformer, the predominant model in natural language processing (NLP), to image classification tasks, achieving competitive performance to top-performing CNNs. With much less inductive bias than CNNs, this ViT architecture adopts a patchify stem where the self-attention is directly computed within non-overlapping local image patches (called visual tokens). Its head merely comprises an FC layer and a softmax classifier, and takes the representation of an extra class token (a learnable embedding) as the input to predict the classification output. The class token interacts with the visual tokens across all multihead self-attention (MSA) layers, resembling transformers in NLP tasks~\cite{transformer,bert}. Subsequent ViTs, such as DeiT~\cite{deit} and PVT~\cite{pvt}, use the same head structure. It is noteworthy that some ViTs~\cite{swin,shufflevit,scalingvit} discard the class token, and adopt a head design similar to GoogLeNet. Specifically, they apply a GAP layer over the feature maps from the last MSA layer to generate head input, achieving slightly better accuracy and better memory efficiency compared to the use of class token. In Table~\ref{table:vits}, we also validate this on ImageNet dataset with Deit-Tiny backbone, showing that the GAP based head shows $0.2\%$ improvement in top-1 accuracy to the class-token based counterpart while our design gets $2.13\%$ improvement. To circumvent the heavy computational cost of self-attention operations in ViTs, a type of conceptually more simple architectures, which is entirely built upon MLP layers repeatedly applied across either the channel-patch dimension or the patch-channel dimension, has been presented~\cite{mlp,mlpmixer,resmlp}. MLPs retain the patchify stem of ViTs, but remove the self-attention component. Regarding the choice of head structure, they adopt the GAP-based design. In a nutshell, the above design instantiations indicate that prevailing CNNs, ViTs and MLPs mostly adopt a similar feature encoding pipeline in their head structures, exploiting global feature dependencies while disregarding local ones. Head structures of this type are simple, and have proven to be effective. However, they are incapable of capturing rich class-specific cues as they coarsely process spatial information about the layout of local features, limiting their feature abstraction abilities and resulting in suboptimal performance for image classification, to some degree. The problem will become more serious to lightweight DNNs constructed for resource-constrained applications~\cite{mobilenets,shufflenet}, as 
illustrated in Table~\ref{table:cnns},~\ref{table:vits} and~\ref{table:mlps}.

There exist some works that improve the head structures of classification DNNs. However, most of them~\cite{bipool,kernelpool,saol,dataaug,resattention} focus on CNN architectures, and very few researches~\cite{query2label,sot} consider ViT architectures, and no research is paid to MLP architectures, to the best of our knowledge. 
~\textit{In this paper, we rethink the feature encoding pipeline in any DNN architecture, and attempt to develop a comparably more universal head structure that can be used to improve the performance of various types of DNN architectures for both multi-class and multi-label image classification scenarios}.

To this end, we present Non-glObal Attentive Head (NOAH), a simple, effective and plug-and-play head structure, which relies on a novel form of dot-product attention called pairwise object category attention (POCA), efficiently generating category-specific attentions at spatial locations for improved feature encoding. Fig.~\ref{fig:noah} depicts an architectural overview of three types of DNN backbones appended with a NOAH. When constructing NOAH, we learn POCAs at local to global scales by a neat combination of feature split, transform and merge operations, taking the feature maps from the last layer of a backbone as the input (for ViT and its variants, we remove the original class token if exists). Specifically, we first split the input into multiple non-overlapping feature groups containing the same number of channels, allowing NOAH to efficiently learn group-wise POCAs in parallel. Upon each feature group, we then take use of a POCA block, which starts from a pair of parallel linear embeddings followed by a mixing operator, to produce a tensor representation of the POCAs at a local scale. Finally, the tensors of local POCAs transformed from all feature groups are merged into a global POCA vector via a simple summation along the spatial dimension, which is fed to a softmax classifier for object category predication.

We first test NOAH with 25 DNN architectures on ImageNet classification benchmark. Specifically, we apply NOAH to 9 CNN backbones, 8 ViT backbones and 8 MLP backbones. For these models, the number of parameters ranges from $1.68$ million to 86.86 million, covering a relatively large range of model size. Despite its simplicity, NOAH usually attains promising performance in terms of both model accuracy and efficiency. When training the models from scratch, the improvement in top-1 accuracy by NOAH ranges from $0.30\%$ to $5.3\%$, without bells and whistles. We showcase the generality of NOAH on three other multi-class and multi-label image classification benchmarks including Market-1501~\cite{market1501}, iNaturalist~\cite{inaturalist} and MS-COCO~\cite{coco} in different training regimes. For instance, NOAH brings $3.6\%$$|$$1.1\%$ mAP gain to ResNet101$|$ViT-Large on MS-COCO. We also conduct a lot of ablation experiments to analyze the key factors and the robustness of NOAH.

\section{Related Work}

\textbf{Backbone architectures for image classification.} CNNs have been the mainstream backbone architectures since the advent of AlexNet~\cite{alexnet} which made breakthrough in 2012 ImageNet classification benchmark. Many follow-up designs, such as VGG~\cite{vgg}, GoogLeNet~\cite{googlenet}, ResNet~\cite{resnet}, DenseNet~\cite{densenet} and ConvNeXt~\cite{convnext}, construct more powerful CNNs by scaling up network depth or width. Some lightweight CNNs, such as MobileNets~\cite{mobilenets,mobilenetv2} and ShuffleNets~\cite{shufflenet,shufflenetv2}, are proposed to meet resource-constrained applications. Unlike the above CNNs that are designed manually, neural architecture search constructs CNNs in an automatic manner~\cite{nas,efficientnet}. Inspired by the great success of transformers~\cite{transformer,bert} in NLP, Dosovitskiy et al. present the first clean transformer architecture ViT~\cite{vit} using a patchify stem. DeiT~\cite{deit} uses a token based distillation strategy to boost the training of ViT models, taking pre-trained large CNN models as the teacher. Many works extend popular architectural practices in constructing CNNs to advance ViT backbone designs, including PVT~\cite{pvt} and Swin~\cite{swin} using pyramid self-attention structures, TNT~\cite{tnt} and T2T~\cite{t2t} using fine-grained patch partition strategies, to name a few. Although ViTs have become a strong alternative to CNNs, they suffer from 
the quadratic complexity of self-attention operations. Mixer~\cite{mlpmixer}, ResMLP~\cite{resmlp} and gMLP~\cite{gmlp} show that replacing self-attention operations by linear operations does not affect the performance, resulting in a class of much simpler DNN architectures entirely built with MLP layers. Subsequent MLPs use extra designs such as shift~\cite{cyclemlp, s2mlp} and permutator~\cite{2vpmlp} modules to strengthen feature communication. In this paper, our focus is to improve the classification performance of these three types of DNNs via rethinking the feature encoding pipeline for designing a universal drop-in head structure. 

\textbf{Methods to improve classification head.} Given a CNN architecture, there exist many works that are directly or indirectly related to design a better classification head. They mainly focus on descriptive methods to pooling~\cite{stopool,bipool,combipool,gsop,gcp,fastgcp,detailpool,liftpool,blurpool,spectralpool,s3pool,lip,kernelpool,gcpjournal,gausspool}, multi-layer/multi-region feature aggregation~\cite{crosslayerpool,saol,dataaug,mcar,harmoniousattention}, parametric learnable embedding~\cite{netvlad,fishernet,g2denet,monet,labelgraph}, attentive regularization~\cite{spatialattention,resattention} and soft augmentation~\cite{softaug}. Many of them suffer from heavy computational cost due to complex structures or optimization methods, and some are tailored to multi-label or fine-grained classification tasks. Comparatively, very few methods are proposed to improve the head structure of ViTs or MLPs. Query2Label~\cite{query2label} uses a multi-layer transformer block to query and pool spatial features extracted from a backbone, which is specialized to multi-label classification tasks. SoT~\cite{sot} extends global covariance pooling popularly used in the heads of CNNs to aggregate the class token and word/visual tokens to build a better head for transformers.

Comparatively, our NOAH is a comparably more universal head structure relied on a new form of dot-product attention called pairwise object category attention learnt from local to global scales, which can be easily used to improve the performance of various types of DNN architectures for both multi-class and multi-label image classification tasks, as validated by extensive experiments. 

\section{Method}

In this section, we describe the motivation, the formulation and the properties of our method.
\subsection{Problem Definition}

\textbf{Global feature encoding.} We begin with reviewing the popular head structure based on global feature encoding, which can be generally defined as:
\begin{equation}
\label{eq:000}
\begin{split}
\vspace{-0.cm}
& \mathbf Z = \mathbf W (GAP(\mathbf F)), \\
& \mathbf P = softmax(\mathbf Z), \\
\end{split}
\vspace{-0.cm}
\end{equation}
where $\mathbf F \in \mathbb R^{H\times W\times C}$ denotes the feature output from the last layer of the backbone of a classification DNN model, having $C$ channels with the spatial size of $H\times W$ (for ViT and MLP models, following~\cite{swin,shufflevit,scalingvit}, we remove the original class token if exists, and reshape the feature output of the backbone into a square shape, then $H$ and $W$ are the number of image patches in column and row, respectively); $GAP$ denotes global average pooling operation; $\mathbf W \in \mathbb R^{M\times C}$ denotes the weight matrix of an FC layer (or a stack of multiple FC layers); $softmax$ denotes the classifier that converts the logit $\mathbf Z \in \mathbb R^M$ into $\mathbf P \in \mathbb R^M$ consisting of $M$ object category probability scores for an input image.

Due to the associative nature of matrix multiplication, we can rewrite the computation of $\mathbf Z$ in Equation~(\ref{eq:000}) into:
\begin{equation}
\vspace{-0.cm}
\begin{split}
\label{eq:001}
& \mathbf Z =  \sum_{i=1}^{H}\sum_{j=1}^{W} \alpha \mathbf Z_{i,j}, \\
& and \ \ \alpha = \frac{1}{HW}, \mathbf Z_{i,j} =  \mathbf W \mathbf F_{i,j},
\end{split}
\vspace{-0.cm}
\end{equation}
where $\mathbf F_{i,j} \in \mathbb R^C$ denotes the feature vector at the spatial location $(i,j)$ of $\mathbf F$, and $\mathbf Z_{i,j} \in \mathbb R^M $ denotes the logit conditioned on $\mathbf F_{i,j}$.

\textbf{Limitation of global feature encoding.} In Equation~(\ref{eq:001}), as $\alpha$ is a constant value, it is clear that all $H\times W$ feature pixels in $\mathbf F$ contribute equally when computing the logit values for all $M$ object categories, leading to a static and coarse encoding of local features. This will result in suboptimal classification performance, especially to lightweight DNN models, as illustrated in Table~\ref{table:cnns},~\ref{table:vits} and~\ref{table:mlps}.

\textbf{Our motivation.} In principle, for an ideal feature encoding pipeline, the importance of feature vectors $\mathbf F_{i,j}, 1\leq i \leq H$ and $1\leq j \leq W$ should be different in terms of both spatial locations and object categories. That is, it is desired to compute the logit scalar $\mathbf Z_m$ for the $m$-th ($1 \leq $m$ \leq M$) object category in the form of:
\begin{equation}
\vspace{-0.cm}
\begin{split}
\label{eq:002}
\mathbf Z_m =  \sum_{i=1}^{H}\sum_{j=1}^{W} \alpha_m^{i,j} \mathbf Z_m^{i,j}, \\
\end{split}
\vspace{-0.cm}\end{equation}
where $\alpha_m^{i,j}$ and $\mathbf Z_m^{i,j}$ denote the weighting scalar and the logit scalar at the spatial location $(i,j)$ for the $m$-th object category, respectively.

Driven by the above analysis, we present Non-glObal Attentive Head (NOAH), a new design of head structure, to improve feature encoding and classification performance. In the design of NOAH, our main motivation is to formulate an efficient dot-product attention mechanism to generate $\alpha_m^{i,j}\mathbf Z_m^{i,j}$, making it location-adaptive. A structural comparison of the popular head and our NOAH is depicted in Fig.~\ref{fig:noah}.

\subsection{Non-glObal Attentive Head}


NOAH relies on a new form of dot-product attention called pairwise object category attention (POCA) which models category-specific attentions at spatial locations. 
We design a neat combination of feature split (two levels), transform, and merge operations, which enables to efficiently learn POCAs at local to global scales. In what follows, we describe the overall structure of NOAH and the formulation of POCA, following the notations in Equation~(\ref{eq:000}).

\textbf{Overall structure of NOAH.} NOAH starts by sequentially splitting the feature input $\mathbf F$ into $N$ non-overlapping even-sized groups ${\mathbf F_{1}, \mathbf F_{2}, ..., \mathbf F_{N}\in \mathbb R^{H\times W\times C/N}}$ along the channel axis (this is called the first-level feature split, which balances feature encoding ability and efficiency, as can be seen from Table~\ref{table:featuresplit}), then uses $N$ POCA blocks with the same structure to these $N$ feature groups in parallel. This results in $N$ tensor representations ${\mathbf Z_{1}, \mathbf Z_{2}, ..., \mathbf Z_{N} \in \mathbb{R}^{H\times W\times M}}$ of local POCAs which are location-adaptive owing to the formulation of local POCAs. 
Next, they are directly merged into a global POCA vector ${\mathbf Z \in \mathbb R^M}$ via a simple summation of them along the spatial dimension.
Finally, the global POCA vector $\mathbf Z$ is fed to a $softmax$ classifier which predicts $M$ object category probability scores for an input image.

\textbf{Formulation of POCA.} Now, it is clear that the formulation of POCA plays a critical role in NOAH. 
POCA is inspired by the dot-product self-attention popularly used in the ViT backbone engineering~\cite{vit}, but it has clear differences in formulation and focus as clarified at the end of this section. Specifically, given the $n$-th feature group $\mathbf F_n\in \mathbb R^{H\times W\times C/N}, 1\leq n \leq N$, our POCA block begins by splitting $\mathbf F_n$ into two disjoint subgroups $\mathbf F_{kn} \in \mathbb R^{H\times W\times C_{k}} $ 
and $\mathbf F_{vn} \in \mathbb R^{H\times W\times C_{v}}$ 
along the channel axis by a split ratio $r$, where $C_{k} = \lfloor {rC/N} \rfloor $ and $C_{v} = \lceil {(1-r)C/N} \rceil$. As a side benefit, this second-level feature split further improves the efficiency of NOAH, yet has slight effect on the model accuracy compared to the counterpart without using it (see Table~\ref{table:featuresplit}). 
Then, we formulate POCA by taking use of a pair of parallel linear embeddings (termed key and value embeddings) and a mixing operator. The key embedding, composed of a $1\times1$ convolutional kernel $\mathbf W_{kn}\in \mathbb R^{1\times 1\times C_k \times M}$ along the channel dimension and a $softmax$ activation function across the spatial dimension, which projects each pixel in $\mathbf F_{kn}$ to a desired object category dimension $M$, producing an attention tensor $\mathbf A_n\in \mathbb R^{H\times W\times M}$ that encodes dense position-specific object category attentions. The value embedding uses another $1\times1$ convolutional kernel $\mathbf W_{vn}\in \mathbb R^{1\times 1\times C_v \times M}$ to $\mathbf F_{vn}$, producing a value tensor $\mathbf V_{n}\in \mathbb R^{H\times W\times M}$ that maintains the same dimensions to the attention tensor $\mathbf A_n$. The final mixing operator makes an interaction between $\mathbf A_n$ and $\mathbf V_n$ via the Hadamard product, generating another tensor $\mathbf Z_n \in \mathbb R^{H\times W\times M}$ capturing the POCAs at a local scale. Mathematically, the POCA block conditioned on the $n$-th feature group $\mathbf F_n$ can be defined as:
\begin{equation}
\vspace{-0.cm}
\label{eq:004}
\begin{split}
 & \mathbf A_n = softmax(\mathbf W_{kn}*\mathbf F_{kn}), \\
 & \mathbf V_n = \mathbf W_{vn}*\mathbf F_{vn}, \\
 & \mathbf Z_n = \mathbf A_n \odot \mathbf V_n,
\end{split}
\vspace{-0.cm}
\end{equation}
where $*$ denotes the convolution operation, and $\odot$ denotes the Hadamard product. According to Equation~(\ref{eq:004}), $\mathbf A_n$ and $\mathbf V_n$ are defined in a location-adaptive manner.

\textbf{Efficiency of NOAH.} Without considering the bias term, a POCA block has $MC$$/$$N$ learnable parameters (Params), and requires $HWMC$$/$$N$$+$$HWM$ multiply-adds (MAdds) plus $HWM$ multiplications for the Hadamard product. The summation operation and the softmax classifier require $HWMN$ Adds and $M$ MAdds, respectively. In total, NOAH has $MC$ Params and requires $HWMC$$+$$2HWMN$$+$$M$ MAdds. 
Because of its simplicity, applying NOAH to replace the original heads of mainstream classification DNNs keeps almost the same model size, as can be seen from extensive experimental results shown in Table~\ref{table:cnns},~\ref{table:vits} and~\ref{table:mlps}. In Table~\ref{table:speed}, we further compare the runtime model speed on both CPU and GPU platforms, showing that the runtime model speed of NOAH is at a similar level to that of the original head (see Table~\ref{table:speed}). 


\textbf{Differences of POCA to self-attention.} Although POCA in NOAH and self-attention in ViT 
both use the dot-product attention mechanism, they are different in formulation: (1) POCA does not use a patchify stem, a query embedding and a position embedding, and does not need to compute the self-attention typically having a quadratic complexity; (2) in a POCA block, the key and value embeddings process two disjoint feature subgroups separately, without a shared input used in the self-attention block; (3) in order to encode dense position-specific object category attentions, the key and value embeddings of a POCA block project each pixel in their corresponding feature subgroups to a desired object category dimension $M$ (Table~\ref{table:poca} and Fig.~\ref{fig:attention} validate its role), and the resulting attention and value tensors are element-wisely mixed via the Hadamard product, which is a key distinction of POCA to the self-attention; (4) for parallel POCA blocks, there is also no feature sharing across them, in sharp contrast to the multihead self-attention (MSA) that shares the same input to all self-attention blocks in the same layer; (5) unlike the MSA that uses a concatenation operation for feature aggregation, NOAH uses a summation operation to merge the output from parallel POCA blocks into a global POCA vector. Besides the above differences in the attention formulation, our NOAH focuses on the classification head design but not the backbone design, and it can be used to improve the performance of different DNN backbones including CNNs,ViTs and MLPs. The above design elements make NOAH suitable as a general head that can be easily used to various types of DNN backbones for both multi-class and multi-label image classification tasks, as extensively validated in the experiments section. Furthermore, NOAH is compatible to DNN backbones that already have advanced attention modules~\cite{residualattention,cbam,se}, yielding promising extra gains as shown in Table~\ref{table:cbam2}.


\section{Experiments}
In this section, we evaluate the performance of NOAH on different image classification benchmarks, study the design and the robustness of NOAH from different aspects.

\subsection{Multi-class Image Classification on ImageNet Dataset}

\textbf{Dataset and experimental setup.} Our basic experiments are conducted 
on the large scale ImageNet dataset~\cite{imagenet}. It consists of over 1.2 million images for training and 50,000 images for validation, including 1,000 object classes. To have a comprehensive evaluation conditioned on the extreme capability of our computational resources, we apply NOAH to a variety of DNN architectures including 9 CNN backbones, 8 ViT backbones and 8 MLP backones, covering a relatively large range of model size (see Table~\ref{table:cnns},~\ref{table:vits},~\ref{table:mlps}). For CNNs, we select backbones from ResNet~\cite{resnet}, MobileNetV2~\cite{mobilenetv2} and ShuffleNetV2~\cite{shufflenetv2} families. For ViTs, we select backbones from DeiT~\cite{deit} and PVT~\cite{pvt} families. For MLPs, we select backbones from Mixer~\cite{mlpmixer} and gMLP~\cite{gmlp} families. In the experiments, we construct our networks by replacing the existing head of each selected DNN architecture by a NOAH, and train our model from the scratch with the same training settings of the baseline model.

For NOAH, we set $N=\{4,8\}$, $r=\{1/2, 1/4,1/8\}$ for different DNNs using
~\textit{an empirical principle}: the smaller the backbone size the larger the $N$, and the larger the $C/N$ the smaller the $r$. We adopt the standard data augmentation to train and evaluate each network. For training, we first resize input images to $256\times256$, then randomly sample $224\times224$ image crops or their horizontal flips. We standardize the cropped images with mean and variance per channel. For evaluation, we use the center crops of the resized images, and report top-1 and top-5 recognition rates on the ImageNet validation set. For fair comparisons, we use the public PyTorch codes of these networks with the exactly same settings to train all baseline models and our models from scratch. Our trained baseline models are either better than or at least on par with the reported ones.~\textit{Experimental details on different DNN backbones are put in supplementary materials}.

\begin{table}[t]
\begin{center}
\vskip -0.in
\caption{Results comparison on ImageNet dataset with CNN backbones. For NOAH, we set: $N=4$, $r=1/2$ in ResNet18; $N=4$, $r=1/8$ in ResNet50, ResNet101 and ResNet152; $N=8$, $r=1/4$ in MobileNetV2 family and ShuffleNetV2. Gains are bolded.}
\label{table:cnns}
\vskip -0.in
\resizebox{0.99\linewidth}{!}{
\begin{tabular}{l|c|c|c}
\toprule
Network & Params(M) & Top-1($\%$) & Top-5($\%$) \\
\hline
ResNet18 & 11.69 & 70.25 & 89.38 \\
+ NOAH  & 11.70  & 71.81 (\textbf{$\uparrow$1.56}) & 90.18 (\textbf{$\uparrow$0.80}) \\
\hline
ResNet50 & 25.56 & 76.23 & 93.01 \\
+ NOAH  & 25.56 & 77.25 (\textbf{$\uparrow$1.02}) & 93.65 (\textbf{$\uparrow$0.64}) \\
\hline
ResNet101 & 44.55  & 77.41 & 93.67 \\
+ NOAH  & 44.56 & 78.22 (\textbf{$\uparrow$0.81}) & 94.13 (\textbf{$\uparrow$0.46}) \\
\hline
ResNet152 & 60.19  & 78.16 & 94.06 \\
+ NOAH  & 60.20 & 78.57 (\textbf{$\uparrow$0.41}) & 94.36 (\textbf{$\uparrow$0.30}) \\
\hline
MobileNetV2 ($1.0\times$) & 3.50 & 72.02 & 90.43 \\
+ NOAH  & 3.52 & 73.35 (\textbf{$\uparrow$1.33}) & 91.13 (\textbf{$\uparrow$0.70}) \\
\hline
MobileNetV2 ($0.75\times$) & 2.64 & 69.65 & 88.99 \\
+ NOAH & 2.65  & 71.44 (\textbf{$\uparrow$1.79}) & 89.87 (\textbf{$\uparrow$0.88}) \\
\hline
MobileNetV2 ($0.5\times$) & 1.97 & 64.30 & 85.21 \\
+ NOAH & 1.98 & 67.44 (\textbf{$\uparrow$3.14}) & 87.11 (\textbf{$\uparrow$1.90}) \\
\hline
MobileNetV2 ($0.35\times$) & 1.68 & 59.62 & 81.79 \\
+ NOAH & 1.69 & 63.40 (\textbf{$\uparrow$3.78}) & 83.91 (\textbf{$\uparrow$2.12}) \\
\hline
ShuffleNetV2  ($1.0\times$) & 2.28 & 69.43 & 88.81 \\
+ NOAH & 2.29 & 70.72 (\textbf{$\uparrow$1.29}) & 89.38 (\textbf{$\uparrow$0.57}) \\
\bottomrule
\end{tabular}
}
\end{center}
\vskip -0.in
\end{table}

\textbf{Basic results on CNNs.} Table~\ref{table:cnns} shows the results comparison on CNNs. Generally, we can see that our NOAH always achieves superior results on all of these CNN backbones than the original head structures that use global feature encoding, maintaining almost the same model size. In terms of top-1 accuracy, we can observe: (1) NOAH can significantly boost the performance of lightweight CNNs (having less than 5 million of learnable parameters), e.g., bringing a top-1 gain of $1.33\sim3.78\%$ and $1.29\%$ for MobileNetV2 family and ShuffleNetV2 ($1.0\times$), respectively; (2) NOAH can generalize well on relatively large CNNs (having about $10\sim45$ million of learnable parameters), improving ResNet50 and ResNet101 by a top-1 gain of $1.02\%$ and $0.81\%$, respectively; (3) NOAH can still bring acceptable performance gains to large CNNs (having over 60 million parameters), e.g., $0.41\%$ top-1 gain to ResNet152. \textit{Note that the performance improvement is obtained under the condition of simply replacing the existing heads of these CNN backbones by the corresponding NOAHs, without bells and whistles both in training and evaluation}. 
The performance improvement of NOAH gradually decreases when the network becomes deeper, larger and more powerful. This is a common experimental trend in deep learning, as large CNN backbones have many more parameters, and tend to have much better learning capacities, compared to smaller CNN backbones. 

\begin{table}[t]
\begin{center}
\vskip -0.in
\caption{Results comparison on ImageNet dataset with ViT backbones. We use a uniform width multiplier $(0.75,0.5)$ to scale down DeiT-Tiny and PVT-Tiny, resembling MobileNetV2~\cite{mobilenetv2}. For NOAH, we set $N = 4$, $r = 1/2$ in all models. Gains are bolded.}
\label{table:vits}
\vskip -0.in
\resizebox{0.99\linewidth}{!}{
\begin{tabular}{l|c|c|c}
\toprule
Network & Params(M) & Top-1($\%$) & Top-5($\%$) \\
\hline
DeiT-Tiny ($1.0\times$) & 5.72 & 72.16 & 91.30 \\
+ GAP & 5.72 & 72.36 ($\uparrow$0.20) & 91.33 ($\uparrow$0.03) \\
+ NOAH & 5.72 & 74.29 (\textbf{$\uparrow$2.13}) & 92.27 (\textbf{$\uparrow$0.97}) \\
\hline
DeiT-Tiny ($0.75\times$) & 3.29 & 62.55 & 85.32 \\
+ NOAH & 3.30 & 66.64 (\textbf{$\uparrow$4.09}) & 87.79 (\textbf{$\uparrow$2.47}) \\
\hline
DeiT-Tiny ($0.5\times$) & 1.53 & 51.36 & 76.79 \\
+ NOAH & 1.54 & 56.66 (\textbf{$\uparrow$5.30}) & 80.41 (\textbf{$\uparrow$3.62}) \\
\hline
DeiT-Small & 22.06 & 79.78 & 94.99 \\
+ NOAH & 22.06 & 80.56 (\textbf{$\uparrow$0.78}) & 95.39 (\textbf{$\uparrow$0.40}) \\
\hline
DeiT-Base & 86.86  & 81.85 & 95.59 \\
+ NOAH & 86.86 & 82.22 (\textbf{$\uparrow$0.37}) & 95.75 (\textbf{$\uparrow$0.16}) \\
\hline
PVT-Tiny ($1.0\times$) & 13.23 & 75.10 & 92.41  \\
+ NOAH & 13.24 & 76.51 (\textbf{$\uparrow$1.41}) & 93.25 (\textbf{$\uparrow$0.84}) \\
\hline
PVT-Tiny ($0.75\times$) & 7.62 & 71.81 & 90.35 \\
+ NOAH & 7.62 & 74.22 (\textbf{$\uparrow$2.41}) & 91.82 (\textbf{$\uparrow$1.47}) \\
\hline
PVT-Tiny ($0.5\times$) & 3.54 & 65.33 & 86.61 \\
+ NOAH & 3.55 & 68.50 (\textbf{$\uparrow$3.17}) & 88.42 (\textbf{$\uparrow$1.81}) \\
\bottomrule
\end{tabular}
}
\end{center}
\vskip -0.in
\end{table}

\textbf{Basic results on ViTs.} From the results in Table~\ref{table:vits}, we can see that the ViT models trained with NOAH show consistently higher accuracy than the baseline models, and the top-1 gain is in the range of $0.37\%\sim5.3\%$. The performance trend of NOAH on ViTs is similar to that on CNNs (shown in Table~\ref{table:cnns}). Compared to the baseline models, our models maintain almost the same model size. 
It is noteworthy that current ViT models are usually much larger than CNN models. For instance, the number of learnable parameters in DeiT-Small and PVT-Tiny is $22.06$ million and $13.23$ million, respectively, although they are named with the keyword ``Small" and ``Tiny". 
In order to better explore the generalization ability of NOAH to more lightweight ViTs, we use a uniform width multiplier ($0.75,0.5$) to scale down the number of feature channels in every building block of DeiT-Small and PVT-Tiny, resembling MobileNetV2~\cite{mobilenetv2}.

\begin{table}[t]
\begin{center}
\vskip -0.in
\caption{Results comparison on ImageNet dataset with MLP backbones. We use a uniform width multiplier $(0.75,0.5)$ to scale down Mixer-Small and gMLP-Tiny, resembling MobileNetV2~\cite{mobilenetv2}. For NOAH, we set $N = 4$, $r = 1/2$ in all models. Gains are bolded.}
\label{table:mlps}
\vskip -0.in
\resizebox{0.99\linewidth}{!}{
\begin{tabular}{l|c|c|c}
\toprule
Network & Params(M) & Top-1($\%$) & Top-5($\%$) \\
\hline
Mixer-Small ($1.0\times$) & 18.53 & 74.18 & 91.56  \\
+ NOAH & 18.54 & 75.09 (\textbf{$\uparrow$0.91}) & 92.22 (\textbf{$\uparrow$0.66}) \\
\hline
Mixer-Small ($0.75\times$) & 10.75 & 71.13 & 90.07 \\
+ NOAH & 10.76 & 72.32 (\textbf{$\uparrow$1.19}) & 90.57 (\textbf{$\uparrow$0.50}) \\
\hline
Mixer-Small ($0.5\times$) & 5.07 & 65.22 & 86.34  \\
+ NOAH & 5.08 & 66.81 (\textbf{$\uparrow$1.59}) & 87.07 (\textbf{$\uparrow$0.73}) \\
\hline
Mixer-Base & 59.88 & 77.14 & 93.02 \\
+ NOAH & 59.88 & 77.49 (\textbf{$\uparrow$0.35}) & 93.27 (\textbf{$\uparrow$0.25}) \\
\hline
gMLP-Tiny ($1.0\times$) & 5.87 & 72.05 & 91.23  \\
+ NOAH & 5.87 & 73.39 (\textbf{$\uparrow$1.34}) & 91.81 (\textbf{$\uparrow$0.58}) \\
\hline
gMLP-Tiny ($0.75\times$) & 3.91 &  65.95 & 87.19  \\
+ NOAH & 3.91 & 67.71 (\textbf{$\uparrow$1.76}) & 88.32 (\textbf{$\uparrow$1.13}) \\
\hline
gMLP-Tiny ($0.5\times$) & 2.41 & 54.99 & 80.02  \\
+ NOAH & 2.41 & 56.89 (\textbf{$\uparrow$1.90}) & 81.00 (\textbf{$\uparrow$0.98}) \\
\hline
gMLP-Small & 19.42 & 79.65 & 94.70  \\
+ NOAH & 19.42 & 79.95 (\textbf{$\uparrow$0.30}) & 94.86 (\textbf{$\uparrow$0.16}) \\
\bottomrule
\end{tabular}
}
\end{center}
\vskip -0.in
\end{table}

\textbf{Basic results on MLPs.} The results comparison on MLPs is given in Table~\ref{table:mlps}. Again, we can find that the MLP models trained with NOAH always achieve better performance than their corresponding baseline models, maintaining almost the same model size. 
Similar to the trend of results on CNNs and ViTs, the performance improvement by NOAH is pronounced when the model size becomes smaller. Just like ViTs, ``Small" MLPs, such as Mixer-Small and gMLP-Small, actually are not small or efficient, compared to CNN counterparts. We also apply the width scaling strategy to Mixer-Small and gMLP-Small as similar to DeiT-Small and PVT-Tiny, and use the resulting more lightweight variants to test the performance of NOAH to thin MLPs.

In Table~\ref{table:speed}, we further provide a comprehensive comparison of the wall clock time of 25 models with NOAH and their corresponding baseline models, showing the efficiency of our method.

\textbf{Results comparison with existing methods.} As discussed in previous sections, there exist many methods that are directly or indirectly related to design a better classification head, given a CNN backbone. In Table~\ref{table:resnet50}, we provide a horizontal performance comparison of our method with a lot of existing methods which report the results for the ResNet50 backbone on ImageNet dataset. It should be noted that these reference methods mostly differ in design focus, optimization and experimental settings, making an apple-to-apple performance comparison not applicable. Under this context, the comparison in Table~\ref{table:resnet50} demonstrates that the performance of NOAH is competitive, to a large extent.

\begin{table}[t]
\begin{center}
\vskip -0.in
\centering
\caption{Horizontal performance comparison of NOAH with existing methods which report the results for the ResNet50 backbone on ImageNet dataset. For NOAH, we set $N=4$, $r=1/8$. The results of the reference methods are collected from the original papers. Since these methods mostly differ in design focus, optimization and experimental settings, an apple-to-apple performance comparison is not applicable. $*$ denotes NOAH with the aggressive training regime~\cite{convnext}. Best results are bolded.}
\label{table:resnet50}
\vskip -0.in
\resizebox{0.99\linewidth}{!}{
\begin{tabular}{l|c|c|c}
\toprule
Network & Params(M) & Top-1($\%$) & Top-5($\%$) \\
\hline
ResNet50 & 25.56 & 76.23 & 93.01 \\
GatedPool~\cite{mixed_gated}  & NA & 77.73 & 93.67 \\
MixedPool~\cite{mixed_gated}  & NA & 77.19 & 93.47 \\
DPP~\cite{detailpool}  & 25.60 & 77.22 & 93.64 \\
BlurPool~\cite{blurpool}  & NA & 77.04 & NA \\
LIP(w/ Bottleneck-128)~\cite{lip}  & 24.70 & 78.19 & 93.96 \\
LIP(w/ Bottleneck-256)~\cite{lip}  & 25.80 & 78.15 & 94.02 \\
GFGP~\cite{gfgp}  & NA & 78.21 & 94.05 \\
Half-GaussPool~\cite{gausspool}  & NA & 78.34 & 94.12 \\
iSP-GaussPool~\cite{gausspool}  & NA & 78.63 & 94.32 \\
3G-Net~\cite{3gnet} & NA & 78.69 & 94.39 \\
GCP~\cite{understandgcp}  & 56.93 & 78.03 & 93.95 \\
SAOL+KD~\cite{saol}  & NA & 77.11 & 93.59 \\
SAOL+KD+Cutmix~\cite{saol}  & NA & 78.85 & 94.24 \\
CSRA~\cite{resattention}  & NA & 75.70 & NA \\
LiftDownPool~\cite{liftpool}  & NA & 77.64 & 93.89 \\
NOAH (ours)  & 25.56 & 77.25 & 93.65 \\
NOAH* (ours) & 25.56 & \textbf{79.17} & \textbf{94.51} \\
\bottomrule
\end{tabular}
}
\end{center}
\vskip -0.in
\end{table}

\begin{table}[t]
\begin{center}
\vskip -0.in
\centering
\caption{Results comparison on MS-COCO dataset. For NOAH, we set: $N=4$, $r=1/8$ in ResNet101; $N = 4$, $r = 1/2$ in ViT-Large. $*$ denotes NOAH with the aggressive training regime~\cite{convnext}. We mainly compare NOAH with existing methods which report the results for the ResNet101 backbone with an image size of $448 \times 448$. Best results are bolded.}
\label{table:mscoco}
\vskip -0.in
\resizebox{0.99\linewidth}{!}{
\begin{tabular}{l|c|c|c|c|c|c|c}
\toprule
\multirow{2}{*}{Methods} & \multicolumn{7}{c}{All($\%$)} \\
\cline{2-8} 
 & mAP & CP & CR & CF1 & OP & OR & OF1 \\ 
\hline
ResNet101 & 80.0 & 82.6 & 68.2 & 74.7 & 85.6 & 72.7 & 78.6 \\
ML-GCN~\cite{mlgcn} & 83.0 & 84.0 & 72.8 & 78.0 & 84.7 & 76.2 & 80.2 \\
A-GCN~\cite{agcn} & 83.1 & 84.7 & 72.3 & 78.0 & 85.6 & 75.5 & 80.3 \\
F-GCN~\cite{fgcn} & 83.2 & 85.4 & 72.4 & 78.3 & 86.0 & 75.7 & 80.5 \\
CMA~\cite{cma} & 83.4 & 83.4 & 72.9 & 77.8 & 86.8 & 76.3 & 80.9 \\
MS-CMA~\cite{cma} & 83.8 & 82.9 & 74.4 & 78.4 & 84.4 & \textbf{77.9} & 81.0 \\
MSRN~\cite{msrn} & 83.4 & \textbf{86.5} & 71.5 & 78.3 & 86.1 & 75.5 & 80.4 \\
MCAR~\cite{mcar} & 83.8 & 85.0 & 72.1 & 78.0 & \textbf{88.0} & 73.9 & 80.3\\
CSRA~\cite{resattention} & 83.5 & 84.1 & 72.5 & 77.9 & 85.6 & 75.7 & 80.3\\
Query2Label~\cite{query2label} & 84.9 & 84.8 & 74.5 & 79.3 & 86.6 & 76.9 & 81.5\\
NOAH (ours) & 83.6 & 85.5 & 71.7 & 78.0 & 86.5 & 75.0 & 80.4\\
NOAH* (ours) & \textbf{85.6} & 86.4 & \textbf{74.8} & \textbf{80.2} & 87.5 & 77.5 & \textbf{82.2} \\
\hline
ViT-Large & 89.2 & \textbf{81.1} & 85.7 & 83.4 & \textbf{81.3} & 87.5 & \textbf{84.3} \\
NOAH (ours) & \textbf{90.3} & 80.3 & \textbf{87.1} & \textbf{83.6} & 79.5 & \textbf{89.1} & 84.0 \\
\bottomrule
\end{tabular}
}
\end{center}
\vskip -0.in
\end{table}

\subsection{Multi-label Image Classification on MS-COCO Dataset}

The above experiments well validate the effectiveness of NOAH to handle multi-class image classification. Next, we conduct comparative experiments on the MS-COCO 2014 dataset~\cite{coco} to explore the performance of NOAH for multi-label image classification.

\textbf{Dataset and experimental setup.} MS-COCO 2014 is a popular dataset for multi-label image classification, which consists of 82,081 images for training and 40,137 images for validation, including 80 object categories. We select ResNet101 (a commonly used benchmarking model on MS-COCO dataset) and ViT-Large~\cite{vit} architectures for experiments, and construct our models by replacing the existing head of each of them by a NOAH. We follow the experimental settings in~\cite{resattention,query2label}, and report mean average precision (mAP), per-category precision (CP), recall (CR), F1-score (CF1), overall precision (OP), recall (OR) and F1-score (OF1) for the overall statistics.
~\textit{Experimental details on different DNN backbones are put in supplementary materials}.

\textbf{Results comparison with existing methods.} Table~\ref{table:mscoco} provides the results comparison on MS-COCO dataset. Note that many existing methods adopt different experimental settings for benchmarking. For a fair comparison, we mainly compare NOAH with existing methods which report the results for the ResNet101 backbone with an image size of $448 \times 448$. It can be seen that NOAH gets an absolute mAP improvement of $3.6\%$ to the ResNet101 model under the standard training regime. With a ResNet101 model pre-trained with the aggressive training regime used in~\cite{convnext}, NOAH even reaches $85.6\%$ mAP, showing the best performance. For much heavier ViT-large model, NOAH still gets $1.1\%$ mAP improvement under the standard training regime.

\subsection{Ablation Studies}

~\textit{To have a deep analysis of NOAH, we conduct a lot of ablative experiments. They are mostly performed on ImageNet dataset, unless otherwise stated}.

\begin{table}[t]
\begin{center}
\vskip -0.in
\centering
\caption{Ablation of NOAH with different settings of $r$ and $N$. Experiments are conducted on ImageNet dataset. Best results are bolded.}
\label{table:rn}
\vskip -0.in
\centering
\resizebox{0.9\linewidth}{!}{
\begin{tabular}{l|c|c|c|c|c}
\toprule
Network & $r$ & $N$ & Params(M) & Top-1($\%$) & Top-5($\%$) \\
\hline
ResNet50 & - & - & 25.56  & 76.23 & 93.01  \\
\hline
\multirow{6}{*}{+ NOAH}
& 1/2 & 4 & 25.56 & 76.84 ($\uparrow$0.61) & 93.32 ($\uparrow$0.31) \\
& 1/4 & 4 & 25.56 & 77.07 ($\uparrow$0.84)& 93.52 ($\uparrow$0.51) \\
& 1/8 & 4 & 25.56 & \textbf{77.25} ($\uparrow$\textbf{1.02}) & \textbf{93.65} ($\uparrow$\textbf{0.64}) \\
\cline{2-6}
& 1/8 & 1 & 25.56 & 76.50 ($\uparrow$0.27)& 93.21 ($\uparrow$0.20) \\
& 1/8 & 2 & 25.56 & 76.77 ($\uparrow$0.54)& 93.36 ($\uparrow$0.35) \\
& 1/8 & 4 & 25.56 & \textbf{77.25} ($\uparrow$\textbf{1.02})& \textbf{93.65} ($\uparrow$\textbf{0.64}) \\
& 1/8 & 8 & 25.57 & 76.94 ($\uparrow$0.71)& 93.34 ($\uparrow$0.33) \\
\bottomrule
\end{tabular}
}
\end{center}
\vskip -0.in
\end{table}

\begin{table}[t]
\begin{center}
\vskip -0.in
\centering
\caption{Ablation of NOAH without vs. with feature split (FS). Experiments are conducted on ImageNet dataset. 
Best results are bolded.}
\label{table:featuresplit}
\vskip -0.in
\centering
\resizebox{0.99\linewidth}{!}{
\begin{tabular}{l|c|c|c|c|c}
\toprule
Network & $1^{st}$ FS  & $2^{nd}$ FS &  Params(M) & Top-1($\%$) & Top-5($\%$) \\
\midrule
ResNet152 & - & - & 60.19 & 78.16 & 94.06 \\
\hline
\multirow{3}{*}{+ NOAH}
& - & - &  74.54 & 78.63 ($\uparrow$0.47) & 94.37 ($\uparrow$0.31) \\
& $\checkmark$ & - & 62.25 &  \textbf{78.77} ($\uparrow$\textbf{0.61})& \textbf{94.47} ($\uparrow$\textbf{0.41}) \\
& $\checkmark$ & $\checkmark$ & 60.20 & 78.57 ($\uparrow$0.41)& 94.36 ($\uparrow$0.30) \\
\hline
ResNet101 & - & - & 44.55 & 77.41 & 93.67 \\
\hline
\multirow{3}{*}{+ NOAH}
& - & - &  58.89 &  78.10 ($\uparrow$0.69)&  94.07 ($\uparrow$0.40) \\
& $\checkmark$ & - & 46.60 & \textbf{78.48} ($\uparrow$\textbf{1.07})& \textbf{94.22} ($\uparrow$\textbf{0.55}) \\
& $\checkmark$ & $\checkmark$ & 44.56 & 78.22 ($\uparrow$0.81)& 94.13 ($\uparrow$0.46) \\
\hline
ResNet50 & - & - & 25.56 & 76.23 & 93.01 \\
\hline
\multirow{3}{*}{+ NOAH}
& - & - & 39.90 & 77.24 ($\uparrow$1.01)&  93.62 ($\uparrow$0.61) \\
& $\checkmark$ & - & 27.61 & \textbf{77.48} ($\uparrow$\textbf{1.25})& \textbf{93.77} ($\uparrow$\textbf{0.76}) \\
& $\checkmark$ & $\checkmark$ & 25.56 & 77.25 ($\uparrow$1.02)& 93.65 ($\uparrow$0.64) \\
\hline
ResNet18 & - & - & 11.69 & 70.25 & 89.38 \\
\hline
\multirow{3}{*}{+ NOAH}
& - & - &  15.28 & \textbf{73.08} ($\uparrow$\textbf{2.83})&  \textbf{91.01} ($\uparrow$\textbf{1.63}) \\
& $\checkmark$ & - & 12.21 & 72.22 ($\uparrow$1.97) & 90.42 ($\uparrow$1.04) \\
& $\checkmark$ & $\checkmark$ & 11.70 & 71.81 ($\uparrow$1.56) & 90.18 ($\uparrow$0.80) \\
\bottomrule
\end{tabular}
}
\end{center}
\vskip -0.in
\end{table}

\textbf{Selection of $N$ and $r$.} Recall that NOAH uses two levels of feature split to learn POCAs at local to global scales. The first-level feature split uses a hyper-parameter $N$ to control the number of POCA blocks in NOAH, and the second-level feature split uses another hyper-parameter $r$ to control the key ratio in each POCA block. Accordingly, our first set of ablative experiments is conducted for the selection of $N$ and $r$. In the experiments, we use ResNet50 as the backbone, and heuristically set $N=4$ to compare different $r$ options first, and then choose the best $r$ to compare different $N$ options. From the results of Table~\ref{table:rn}, we can find that all settings improve the model accuracy while maintaining almost the same model size. Comparatively, the setting of $N=4$ and $r=1/8$ is the best, so we choose it for ResNet50. For the other 24 backbones, we simply adjust this setting via an empirical principle: the smaller the backbone size 
the larger the $N$, and the larger the $C/N$ the smaller the $r$, without tuning $N$ and $r$ network by network for better model accuracy. 

\textbf{Role of the feature split.} In Table~\ref{table:featuresplit}, a comparison of NOAH based models trained without vs. with feature split operations is given. We can see that: (1) when removing two levels of feature split, NOAH shows better results on small ResNet18 backbone but slightly worse results on larger ResNet50/ResNet101 backbone. However, the model size is significantly increased compared to the baseline model, e.g., the number of parameters for ResNet50 with NOAH increases from 25.56 million to 39.9 million; 
(2) the standard NOAH with two levels of feature split can well balance model accuracy and efficiency; (3) 
removing the second-level feature split from the standard NOAH always gets more accurate models with few extra parameters.

\begin{table}[t]
\begin{center}
\vskip -0.in
\centering
\caption{Ablation of the POCA block with a softmax function along different feature dimensions to obtain attention tensor $\mathbf A_{n}$. Experiments are conducted on ImageNet dataset. Best results are bolded.}
\label{table:poca}
\vskip -0.in
\centering
\resizebox{0.99\linewidth}{!}{
\begin{tabular}{l|l|c|c|c}
\toprule
Network & Normalizing POCAs  & Params(M)  & Top-1($\%$) & Top-5($\%$) \\
\midrule
ResNet50 & - & 25.56 & 76.23 & 93.01 \\
\hline
\multirow{2}{*}{$\text{+ NOAH}$}
 & Along channel & 25.56 & 75.06 ($\downarrow$1.17) & 92.50 ($\downarrow$0.51)  \\
 & Along spatial (ours) & 25.56 & \textbf{77.25} ($\uparrow$\textbf{1.02}) & \textbf{93.65} ($\uparrow$\textbf{0.64}) \\
\bottomrule
\end{tabular}
}
\end{center}
\vskip -0.in
\end{table}

\begin{table}[t]
\begin{center}
\vskip -0.in
\centering
\caption{Ablation of NOAH using different operations to merge $N$ local POCA tensors into a global POCA vector. Experiments are conducted on ImageNet dataset. Best results are bolded.}
\label{table:sum}
\vskip -0.in
\centering
\resizebox{0.99\linewidth}{!}{
\begin{tabular}{l|l|c|c|c}
\toprule
Network & Merge Local POCAs & Params(M) & Top-1($\%$) & Top-5($\%$) \\
\midrule
ResNet50 & - & 25.56 & 76.23 & 93.01 \\
\hline
\multirow{3}{*}{$\text{+ NOAH}$}
 & Max pooling & 25.56 & 68.14 ($\downarrow$8.09) & 88.06 ($\downarrow$4.95) \\
 & Average pooling & 25.56 & 70.56 ($\downarrow$5.67) & 89.79 ($\downarrow$3.22) \\
 & Summation (ours)  & 25.56 & \textbf{77.25} ($\uparrow$\textbf{1.02}) & \textbf{93.65} ($\uparrow$\textbf{0.64}) \\
\bottomrule
\end{tabular}
}
\end{center}
\vskip -0.in
\end{table}

\textbf{Role of the summation for merging local POCAs.} In the POCA block, we use a $softmax$ function along the spatial dimension to generate the attention tensor $\mathbf A_n$, which is an important design of NOAH to learn POCAs at a local scale. Another natural choice is to apply the $softmax$ function along the channel dimension. In Table~\ref{table:poca}, we study these two choices using ResNet50 as the backbone and the setting of $N=4$ and $r=1/8$. Surprisingly, the $softmax$ function along the channel dimension leads to $1.17\%$ top-1 accuracy drop against the baseline, but our design brings $1.02\%$ improvement. This suggests that explicitly modeling dense spatial feature dependencies by producing a spatial attention map for each object category is essential for the POCA block. Moreover, we use a summation operation along the spatial dimension to merge $N$ local POCA tensors into a global POCA vector. 
In Table~\ref{table:sum}, we further compare the summation with max pooling and average pooling, showing its advantage.

\begin{table}[t]
\begin{center}
\vskip -0.in
\caption{Ablation of comparing NOAH with different attention designs for the POCA block. Experiments are conducted on ImageNet dataset. Best results are bolded.}
\label{table:7pocas}
\vskip 0.05in
\resizebox{0.98\linewidth}{!}{
\begin{tabular}{l|l|c|c|c}
\toprule
Network & POCA Variants & Params(M) & Top-1($\%$) & Top-5($\%$) \\
\hline
ResNet18 & - & 11.69 & 70.25 & 89.38 \\
\hline
\multirow{5}{*}{$\text{+ NOAH}$}
 & Standard POCA & 11.70 & \textbf{71.81} ($\uparrow$\textbf{1.56})  & \textbf{90.18} ($\uparrow$\textbf{0.80}) \\
 & POCA variant (a) & 11.96 & 71.37 ($\uparrow$1.12) & 90.01 ($\uparrow$0.63) \\
 & POCA variant (b)   & 11.76 & 70.42 ($\uparrow$0.17) & 89.59 ($\uparrow$0.21) \\
 & POCA variant (c) & 11.70 & 70.68 ($\uparrow$0.43) & 89.64 ($\uparrow$0.26) \\
 & POCA variant (d) & 15.79 & 71.57 ($\uparrow$1.32) & 90.05 ($\uparrow$0.67) \\
 & POCA variant (e) & 11.69 & 70.26 ($\uparrow$0.01) & 89.41 ($\uparrow$0.03) \\
 & POCA variant (f) & 11.73 & 71.28 ($\uparrow$1.03) & 89.97 ($\uparrow$0.59) \\
\bottomrule
\end{tabular}
}
\end{center}
\vskip -0.in
\end{table}

\begin{table}[t]
\begin{center}
\vskip -0.in
\caption{Ablation of comparing NOAH with the self-attention based head. Experiments are conducted on ImageNet dataset. Best results are bolded.}
\label{table:vithead}
\vskip 0.05in
\resizebox{0.98\linewidth}{!}{
\begin{tabular}{l|c|c|c}
\toprule
Network & Params(M) & Top-1($\%$) & Top-5($\%$) \\
\hline
ResNet50 & 25.56 & 76.23 & 93.01 \\
\hline
+ NOAH	& 25.56	& \textbf{77.25}	&\textbf{93.65} \\
+ Self-attention based head  &27.04	&75.61	&92.63 \\
\bottomrule
\end{tabular}
}
\end{center}
\vskip -0.in
\end{table}

\textbf{Variant designs for the POCA block.} Clearly, the POCA block plays the key role in NOAH. For a better understanding of our proposed attention block for POCA, we also perform ablative experiments with ResNet18 as the backbone to compare our standard design (using the setting of $N=4$ and $r=1/2$) for the POCA block with the following 6 variant designs: (a) for the desired channel number of key embedding $\mathbf W_{kn}$ in each POCA block, reducing it from $1000$ (the number of object categories $M$) to $512$, and adding an extra FC layer to project the global POCA feature to the object category dimension; (b) for the desired channel number of key embedding $\mathbf W_{kn}$ in each POCA block, reducing it from $1000$ to $128$, applying the summation operation to $N$ POCA blocks separately, and adding an extra FC layer to project the concatenated feature of $N$ POCA vectors to the object category dimension $M$; (c) for the activation function, replacing the $softmax$ by the $sigmoid$; (d) for the key embedding $\mathbf{W}_{kn}$ and the value embedding $\mathbf{W}_{vn}$, replacing $1\times1$ convolutional kernels by $3\times3$ convolutional kernels; (e) for the key embedding $\mathbf{W}_{kn}$, removing the second-level feature split, and generating a single (instead of $1000$, that is, one unique spatial attention matrix per object category) spatial attention $\mathbf A_n\in \mathbb{R}^{H\times W\times 1}$ which is shared to the value tensor $\mathbf V_n\in \mathbb{R}^{H\times W\times M}$ along the object category dimension $M$; (f) adding an extra linear ($1\times1$ convolutional) layer before the key embedding $\mathbf{W}_{kn}$, and also adding an extra linear ($1\times1$ convolutional) layer before the value embedding $\mathbf{W}_{kn}$, whose output features have the same dimensions to the input features, respectively. Table~\ref{table:7pocas} shows the results, from which we can see that all these 7 attention designs for the POCA block can improve model accuracy. Comparatively, our proposed design is the best, bringing $1.56\%$ top-1 accuracy improvement to the baseline while maintaining almost the same model size. Surprisingly, replacing $1\times1$ convolutional kernels by $3\times3$ convolutional kernels does not lead to improved model accuracy although it introduces more pixels (that encode a larger spatial field) to compute each local POCA. Furthermore, reducing the desired channel number of key embedding $\mathbf W_{kn}$ in each POCA block from the number of object categories $M$ to a smaller value decreases the performance improvement of NOAH, which shows the importance of our POCA design to some extent.

\textbf{Comparison of NOAH to the self-attention based head.} Using a standard self-attention block as the final pooling layer can be another baseline for comparison. We also perform a set of experiments on ImageNet dataset to compare our NOAH with such kind of baseline. In the experiments, we use a standard self-attention block as the classification head, which consists of a self-attention layer and an FC layer. The self-attention layer takes each pixel as a token. The input and output channel numbers of the FC layer are set to 512 and 1000, respectively. We add a global average pooling layer before the FC layer for comparison, which is called self-attention based head. The results are shown in Table~\ref{table:vithead}, from which we can see that NOAH performs better than this counterpart using a self-attention block in its classification head.

\begin{table}[t]
\begin{center}
\vskip -0.in
\caption{Ablation of comparing NOAH with popular attention modules tailored to the DNN backbone design. Experiments are conducted on ImageNet dataset with the ResNet18 backbone. Best results are bolded.}
\vskip -0.in
\label{table:cbam1}
\resizebox{0.9\linewidth}{!}{
\begin{tabular}{l|c|c|c}
\hline
Network & Params(M) &  Top-1(\%) & Top-5(\%)\\
\hline
ResNet18 & 11.69 & 70.25 & 89.38 \\
\hline
+ SE~\cite{se} & 11.78 & 70.98 ($\uparrow$0.73) & 90.03 ($\uparrow$0.65)\\
+ CBAM~\cite{cbam} & 11.78 & 71.01 ($\uparrow$0.76) & 89.85 ($\uparrow$0.47)\\
+ ECA~\cite{eca} & 11.69 & 70.60 ($\uparrow$0.35) & 89.68 ($\uparrow$0.30)\\
+ CGC~\cite{cgc} & 11.69 & 71.60 ($\uparrow$1.35) & 90.35 ($\uparrow$0.97)\\
+ WeightNet~\cite{weightnet} & 11.93 & 71.56 ($\uparrow$1.31) & 90.38 ($\uparrow$1.00) \\
+ WE~\cite{we} & 11.90 & 71.00 ($\uparrow$0.75) & 90.00 ($\uparrow$0.62)\\
\hline
+ NOAH & 11.70 &  71.81 ($\uparrow$1.56) & 90.18 ($\uparrow$0.80)\\
+ NOAH \& CBAM & 11.78 & \textbf{72.17} ($\uparrow$\textbf{1.92}) & \textbf{90.50} ($\uparrow$\textbf{1.12})\\
\hline
\end{tabular}
}
\end{center}
\vskip -0. in
\end{table}

\begin{table}[t]
\begin{center}
\caption{Ablation of comparing NOAH with popular attention modules tailored to the DNN backbone design. Experiments are conducted on ImageNet dataset with the ResNet50 backbone. Best results are bolded.}
\vskip -0. in
\label{table:cbam2}
\resizebox{0.9\linewidth}{!}{
\begin{tabular}{l|c|c|c}
\hline
Network & Params(M) &  Top-1(\%) & Top-5(\%)\\
\hline
ResNet50 & 25.56 & 76.23 & 93.01 \\
\hline
+ SE~\cite{se} & 28.07 & 77.37 ($\uparrow$1.14)& 93.63 ($\uparrow$0.62)\\
+ CBAM~\cite{cbam} & 28.07 & 77.46 ($\uparrow$1.23)& 93.59 ($\uparrow$0.58)\\
+ ECA~\cite{eca} & 25.56 & 77.34 ($\uparrow$1.11)& 93.64 ($\uparrow$0.63)\\
+ CGC~\cite{cgc} & 25.59 & 76.79 ($\uparrow$0.56)& 93.37 ($\uparrow$0.36)\\
+ WeightNet~\cite{weightnet} & 30.38 & 77.43 ($\uparrow$1.20)& 93.65 ($\uparrow$0.64)\\
+ WE~\cite{we} & 28.10 & 77.10 ($\uparrow$0.87)& 93.50 ($\uparrow$0.49)\\
\hline
+ NOAH & 25.56 & 77.25 ($\uparrow$1.02) & 93.65 ($\uparrow$0.64) \\
+ NOAH \& CBAM & 28.08 & \textbf{77.80} ($\uparrow$\textbf{1.57}) & \textbf{93.90} ($\uparrow$\textbf{0.89})\\
\hline
\end{tabular}
}
\end{center}
\vskip -0. in
\end{table}

\textbf{Combination of head attention and backbone attention.} Note that substantial works have been presented to the backbone design of DNN architectures. One promising way to improve the performance of DNN models is to modify the backbone by using advanced attention modules~\cite{se,cbam,eca,we,weightnet,cgc}. Comparatively, our NOAH is designed as a universal head structure, which can be readily used to improve the performance of various types of DNN architectures for different image classification tasks. Because of different focuses and formulations, NOAH is compatible to DNN backbones already having advanced attention modules. In Table~\ref{table:cbam1} and~\ref{table:cbam2}, we provide two sets of ablative experiments to compare the performance of NOAH and popular attention modules tailored to the DNN backbone design, and also explore the potential of combing NOAH with CBAM~\cite{cbam} (which jointly uses channel attention and spatial attention modules to the backbone). We select ResNet18 and ResNet50 as two backbones for experiments, and train the models on ImageNet dataset using the standard training settings (the same to those for Table~\ref{table:cnns}). From the results of Table~\ref{table:cbam1} and~\ref{table:cbam2}, we can see that NOAH gets very competitive performance, and the best models are obtained on both backbones when combining NOAH with CBAM.

\begin{table}[t]
\begin{center}
\vskip -0.in
\centering
\caption{Ablation of NOAH on fine-grained classification dataset iNaturalist having 5089 categories. Best results are bolded.} 
\label{table:inaturalist}
\vskip -0.in
\resizebox{0.99\linewidth}{!}{
\begin{tabular}{l|c|c|c|c|c}
\toprule
Network & Params(M) & Top-1($\%$) & Network & Params(M) & Top-1($\%$)  \\
\hline
ResNet18 & 13.79 & 50.59  & ResNet50 & 33.94 & 58.20 \\
+ NOAH & 13.82 & \textbf{52.37} ($\uparrow$\textbf{1.78}) & + NOAH & 33.97 & \textbf{59.88} ($\uparrow$\textbf{1.68})\\
\bottomrule
\end{tabular}
}
\end{center}
\vskip -0.in
\end{table}

\begin{table}[t]
\begin{center}
\vskip -0.in
\centering
\caption{Ablation of NOAH on person re-identification dataset Market-1501 having 1501 identities. Best results are bolded.}
\label{table:market1501}
\vskip -0.in
\resizebox{.85\linewidth}{!}{
\begin{tabular}{l|c|c|c|c}
\toprule
\multirow{2}{*}{Network} & \multicolumn{2}{c|}{Training from scratch} &  \multicolumn{2}{c}{Fine-tuning} \\
\cline{2-5}
 & Top-1($\%$) & mAP($\%$) & Top-1($\%$) & mAP($\%$) \\
\midrule
ResNet50 & 91.5 & 78.2 & 92.2 & 79.9 \\
+ NOAH & \textbf{93.5($\uparrow$2.0)} & \textbf{81.8($\uparrow$3.6)} & \textbf{93.7($\uparrow$1.5)} & \textbf{82.3($\uparrow$2.4)} \\
\bottomrule
\end{tabular}
}
\end{center}
\vskip -0.in
\end{table}

\textbf{Fine-grained Image Classification on iNaturalist Dataset.} We also explore the generalization ability of NOAH to handle fine-grained image classification task using iNaturalist 2017 dataset~\cite{inaturalist}, which consists of 579,184 images for training and 95,986 images for validation, including 5,089 classes. Compared to ImageNet dataset, iNaturalist 2017 dataset has significantly more image classes and imbalanced class distribution. In the experiments, we use ResNet18 and ResNet50 as the backbones (the setting of $N$ and $r$ is the same to that in Table~\ref{table:cnns}) and train them with the standard training from scratch regime. The results of Table~\ref{table:inaturalist} show that NOAH attains promising performance again, bringing $1.78\%$$|$$1.68\%$ top-1 accuracy improvement to ResNet18$|$ResNet50 backbone.~\textit{Experimental details on different DNN backbones are put in supplementary materials}.

\textbf{Person Re-identification on Market-1501 Dataset.} Next, we perform ablative experiments on the popular person re-identification dataset Market-1501~\cite{market1501}, which contains 750 and 751 identities for training and testing. In the experiments, we follow the common settings on Market-1501 dataset, and consider two training regimes: the standard from-scratch training and the fine-tuning. Table~\ref{table:market1501} summarizes the results, showing that NOAH consistently achieves large margins against the baseline models trained in these two different training regimes, in terms of both top-1 accuracy and mAP. Comparatively, NOAH shows better results with the fine-tuning than the from-scratch training. However, NOAH gets higher accuracy margins for the from-scratch training than the fine-tuning, and the margin gaps are $0.5\%$ for top-1 accuracy ($2.0\%$ vs. $1.5\%$) and $1.2\%$ for mAP ($3.6\%$ vs. $2.4\%$), respectively.~\textit{Experimental details for two training regimes are put in supplementary materials}.

\textbf{Effect of the aggressive training regime.} Recent work~\cite{convnext} shows that, compared to the standard training regime, much more accurate models can be attained when properly using strong training augmentations like increasing the number of training epochs by multiple times, a linear warmup with tens of epochs, a cosine weight decay, a combination of data augmentation methods, multiple regularization strategies to alleviate overfitting, etc. We also study the generalization ability of NOAH in this aggressive from-scratch model training regime. Experiments are performed on ImageNet dataset with ResNet50 backbone. Table~\ref{table:aggressive} shows the results. We can see that NOAH performs well in this aggressive training regime, no matter using feature split operations or not. With two levels of feature split, NOAH brings $0.73\%$ top-1 gain to the baseline model. Without the second-level feature split, the ResNet50 model trained with NOAH reaches $79.32\%$ top-1 accuracy, showing $0.88\%$ gain to the baseline model.

\begin{table}[t]
\begin{center}
\vskip -0.in
\caption{Ablation of NOAH with the aggressive from-scratch training regime used in~\cite{convnext}. Experiments are conducted on ImageNet dataset. Best results are bolded.}
\label{table:aggressive}
\vskip -0.in
\resizebox{0.92\linewidth}{!}{
\begin{tabular}{l|c|c|c}
\toprule
Network & Params(M) & Top-1($\%$) & Top-5($\%$) \\
\midrule
ResNet50 & 25.56 & 78.44 & 94.24 \\ \hline
+ NOAH & 25.56 &  79.17 ($\uparrow$0.73) & 94.51 ($\uparrow$0.27) \\
+ NOAH (w/o $2^{nd}$ FS) & 27.61 & \textbf{79.32} ($\uparrow$\textbf{0.88}) & \textbf{94.60} ($\uparrow$\textbf{0.36}) \\
+ NOAH (no FS) & 39.90 & 79.00 ($\uparrow$0.56) & 94.33 ($\uparrow$0.09) \\
\bottomrule
\end{tabular}
}
\end{center}
\vskip -0.0in
\end{table}

\subsection{Attention Visualizations}

Recall that our NOAH leverages a neat combination of feature split (two levels), transform (a pair of linear ebmeddings) and merge (summation) operations to learn local-to-global POCAs in a group-wise manner. To have a better understanding of this parallel POCA learning mechanism, it is necessary to study the learnt POCA features. To this end, we conduct visualization experiments to study NOAH from two aspects: (1) analyzing learnt attention tensors between different POCA blocks for the ground truth object category; (2) analyzing learnt attention tensor of the same POCA block for different object categories.

\begin{figure*}[htbp]
\centering
\subfloat[]{
\begin{minipage}[t]{1.0\linewidth}
\centering
\includegraphics[width=0.95\linewidth]{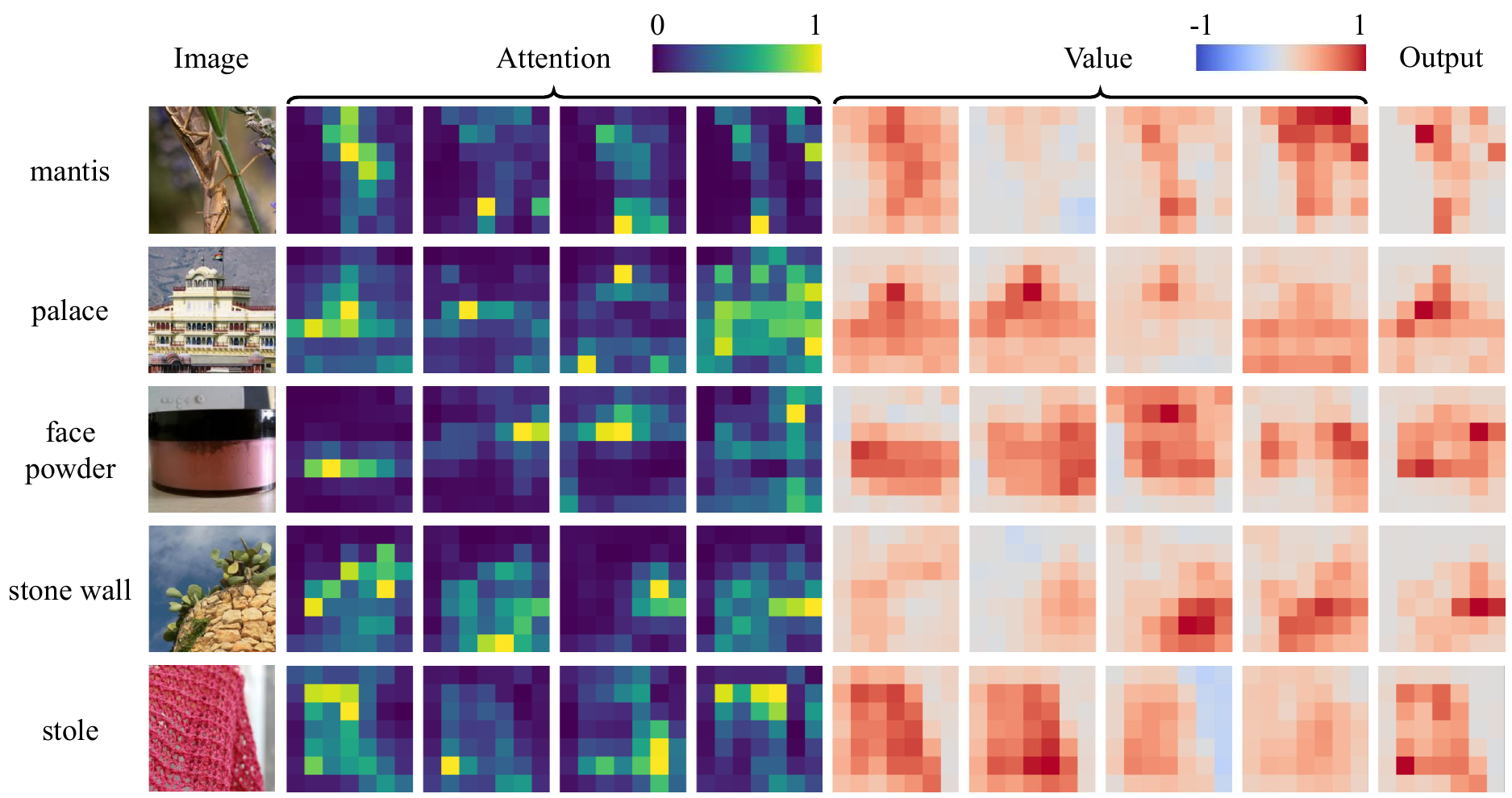}
\end{minipage}%
}
\newline
\subfloat[]{
\begin{minipage}[t]{1.0\linewidth}
\centering
\includegraphics[width=0.98\linewidth]{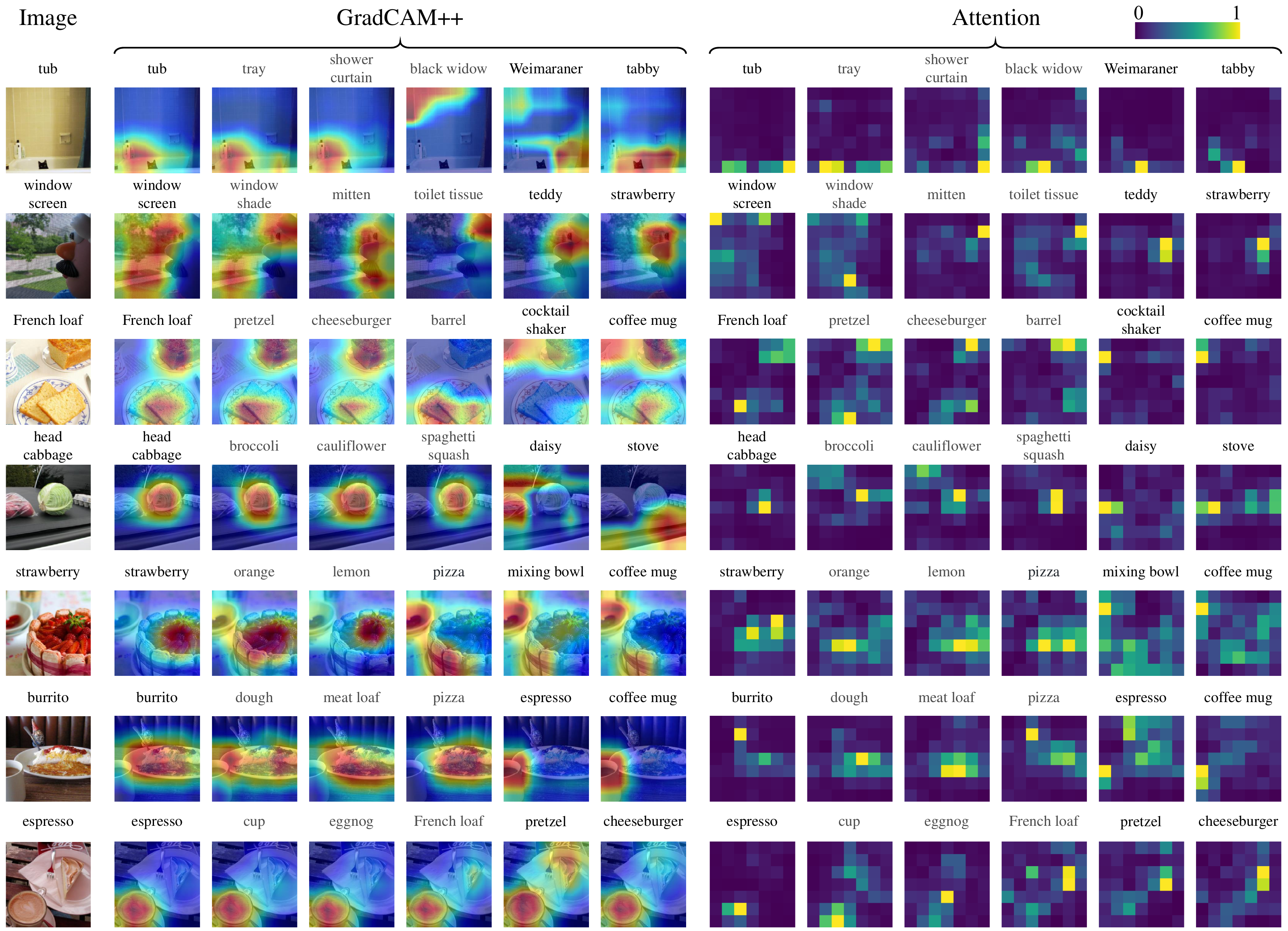}
\end{minipage}%
}
\centering
\vskip -0.in
   \caption{Visualizations of NOAH: (a) illustrative visualizations of learnt attention tensors between different POCA blocks for the ground truth object category, and (b) illustrative visualizations of learnt attention tensors of the same POCA block for different object categories. We use the well-trained ResNet18 model with NOAH and image samples in the ImageNet validation set. For comparison, in (b) we also present the visualization results obtained from the well-trained ResNet18 model with the GAP-based head using Grad-CAM++~\cite{gradcam} for the corresponding object categories.}
\label{fig:attention}
\vskip -0.in
\end{figure*}

First, we use the well-trained ResNet18 model with NOAH (in Table~\ref{table:cnns}) to analyze the learnt attention tensor $\mathbf A_n\in \mathbb{R}^{H\times W\times M}$, the learnt value tensor $\mathbf V_{n}\in \mathbb{R}^{H\times W\times M}$, and the learnt local POCA tensor $\mathbf Z_n \in \mathbb{R}^{H\times W\times M}$ for each of 4 POCA blocks. Given any image sample in the ImageNet validation set, for visualization, we select the spatial attention/value channel (that corresponds to the ground truth object category, and is normalized into $[0,1]/[-1,1]$ for visualization) of $\mathbf{A}_n$/$\mathbf{V}_n$ for each of 4 POCA blocks, and the summation output (that corresponds to the ground truth object category, and is normalized into $[-1,1]$ for visualization) of the local POCA channels of $\mathbf{Z}_n$ tensors generated by 4 POCA blocks, respectively. Illustrative results are shown in Fig.~\ref{fig:attention}. Here, for the ResNet18 trained with NOAH, $N = 4$ and $r = 1/2$. We can observe that parallel POCA blocks tend to learn varying spatial object category attention distributions, which are complementary to each other by visualization examples, showing their capability to capture rich spatial context cues to some degree.

\begin{table*}[t]
\begin{center}
\vskip -0.in
\caption{Speed comparison with 25 different CNN, ViT and MLP backbones trained on ImageNet dataset . We compare the runtime inference speed (Frames Per Second, FPS) of our trained models with NOAHs vs. baseline models. All models are tested on an NVIDIA TITAN X GPU (with batch size 200) and a single core of Intel E5-2683 v3 CPU (with batch size 1), separately. The input image size is 224$\times$224.}
\label{table:speed}
\vskip -0.in
\resizebox{0.93\linewidth}{!}{
\begin{tabular}{l|c|c|c|c|c|c}
\toprule
\multirow{2}{*}{Network} & \multicolumn{3}{c|}{Vanilla} & \multicolumn{3}{c}{NOAH} \\
\cline{2-7}
& Params(M) & Speed on GPU(FPS) & Speed on CPU(FPS) & Params(M) & Speed on GPU(FPS) & Speed on CPU(FPS)  \\
\hline
ResNet18 & 11.69 & 2394.00 & 12.98 & 11.70 & 2117.88 & 12.20 \\ \hline 
ResNet50 & 25.56 & 674.44 & 5.46 & 25.56 & 641.36 & 5.25 \\ \hline 
ResNet101 & 44.55 & 398.98 & 3.20 & 44.56 & 388.38 & 3.08 \\ \hline 
ResNet152 & 60.19 & 275.08 & 2.15 & 60.20 & 269.70 & 2.09 \\ \hline 
MobileNetV2 ($1.0\times$) & 3.50 & 1522.62 & 17.38 & 3.52 & 1361.38 & 16.06 \\ \hline 
MobileNetV2 ($0.75\times$) & 2.64 & 1762.84 & 20.31 & 2.65 & 1565.28 & 17.32 \\ \hline 
MobileNetV2 ($0.5\times$) & 1.97 & 2829.84 & 28.43 & 1.98 & 2361.18 & 22.88 \\ \hline 
MobileNetV2 ($0.35\times$) & 1.68 & 3576.58 & 36.09 & 1.69 & 2850.02 & 28.60 \\ \hline 
ShuffleNetV2 ($1.0\times$) & 2.28 & 4241.94 & 34.36 & 2.29 & 3312.82 & 29.21 \\ \hline 
DeiT-Tiny ($1.0\times$) & 5.72 & 1246.51 & 15.76 & 5.72 & 1039.72 & 14.55 \\ \hline 
DeiT-Tiny ($0.75\times$) & 3.29 & 1606.39 & 23.52 & 3.30 & 1271.78 & 20.74 \\ \hline 
DeiT-Tiny ($0.5\times$) & 1.53 & 2169.99 & 34.51 & 1.54 & 1644.79 & 29.30 \\ \hline 
DeiT-Small & 22.06 & 525.86 & 5.71 & 22.06 & 456.79 & 5.48 \\ \hline 
DeiT-Base & 86.86 & 172.59 & 1.64 & 86.86 & 164.16 & 1.59 \\ \hline 
PVT-Tiny ($1.0\times$) & 13.23 & 619.85 & 9.13 & 13.24 & 603.67 & 8.82 \\ \hline 
PVT-Tiny ($0.75\times$) & 7.62 & 804.17 & 14.35 & 7.62 & 766.34 & 13.20 \\ \hline 
PVT-Tiny ($0.5\times$) & 3.54 & 1088.50 & 23.75 & 3.55 & 1035.93 & 20.47 \\ \hline 
Mixer-Small ($1.0\times$) & 18.53 & 733.96 & 7.18 & 18.54 & 674.54 & 6.81 \\ \hline 
Mixer-Small ($0.75\times$) & 10.75 & 1050.20 & 11.25 & 10.76 & 929.10 & 10.57 \\ \hline 
Mixer-Small ($0.5\times$) & 5.07 & 2037.45 & 23.88 & 5.08 & 1615.76 & 20.62 \\ \hline 
Mixer-Base & 59.88 & 236.49 & 2.13 & 59.88 & 229.55 & 2.05 \\ \hline 
gMLP-Tiny ($1.0\times$) & 5.87 & 776.58 & 13.65 & 5.87 & 712.21 & 12.31 \\ \hline 
gMLP-Tiny ($0.75\times$) & 3.91 & 987.63 & 19.33 & 3.91 & 882.56 & 17.29 \\ \hline 
gMLP-Tiny ($0.5\times$) & 2.41 & 1343.41 & 26.95 & 2.41 & 1140.75 & 23.10 \\ \hline 
gMLP-Small & 19.42 & 381.12 & 5.57 & 19.42 & 368.22 & 5.28 \\ 
\bottomrule
\end{tabular}
}
\end{center}
\vskip -0.in
\end{table*}

Then, we use the well-trained ResNet18 model with NOAH (in Table~\ref{table:cnns}) to analyze the learnt attention tensor $\mathbf{A}_{n}$ for different object categories of the same POCA block. Given any image sample in the ImageNet validation set, for visualization, we select the spatial attention channels (that correspond to the ground truth object category, and representative object categories mostly similar or related to the ground truth) of $\mathbf{A}_{n}$ for the first POCA (i.e., $\mathbf{A}_{1}$) block. Illustrative attention channels are normalized to $[0,1]$ for visualization. For comparison, we also provide the visualization results obtained from the well-trained ResNet18 model with the GAP-based head using Grad-CAM++~\cite{gradcam} for the corresponding object categories. From visualization examples shown in Fig.~\ref{fig:attention}, we can observe that the POCA block can learn different spatial attentions for each object category, and tends to learn roughly similar spatial attention distribution between similar object categories, indicating its capability to capture latent relationship between object categories.

\subsection{Runtime Model Speed}

The wall-clock time at inference is critical to deploy a well-trained model in real applications. We also perform comprehensive experiments on ImageNet dataset to compare the runtime inference speed of the models trained with NOAH and the baseline models using global feature encoding in their head structures. Specifically, we use an NVIDIA TITAN X GPU (with batch size 200) and a single core of Intel E5-2683 v3 CPU (with batch size 1) to test and compare all 25 pairs of the NOAH based model and the baseline model, including 9 CNN pairs, 8 ViT pairs and 8 MLP pairs. Detailed results are shown in Table~\ref{table:speed}. We can observe that: (1) both on GPU and CPU, the models trained with NOAH in general show the relatively slower runtime speed than the baselines, but the extra latency by NOAH is decent ($1.96\%\sim24.20\%$ on GPU, and $2.79\%\sim20.75\%$ on CPU) considering the accuracy improvement; (2) the extra latency by NOAH gradually increases when the network size becomes smaller; (3) similar model speed trends are demonstrated on different CNN, ViT and MLP architectures. It is worth mentioning that all these results are obtained under the condition that the NOAH based models maintain almost the same model size to the respective baseline models (the number of parameters ranges from $1.68$ million to $86.86$ million). 

\subsection{Limitations of NOAH}

Clearly, the above experiments have well validated that NOAH has favorable abilities to improve the representation learning of a variety of CNN, ViT and MLP architectures for different image classification tasks. The main limitation of NOAH lies in the generalization to dense downstream tasks, mainly due to different training paradigms between image classification and dense downstream tasks. For example, when training a deep model for object detection, it typically starts with removing the head of a pre-trained classification network, then appends two new head structures (one for region-based classification, and the other for object localization) to the pre-trained backbone, and finally fine-tunes them on the given dataset while keeping either all layers or some particular layers of the pre-trained backbone fixed. Such a fundamental training paradigm difference makes NOAH cannot be directly used to dense downstream tasks. Our preliminary test of merely using the backbone pre-trained with NOAH to object detection dataset MS-COCO only brings about $0.3\%$ mAP gain. Besides, restricted by our available computational resources, the potential of applying NOAH to super-large DNN architectures (particularly ViTs and MLPs) is not explored. We hope we can explore it in the future.

\section{Conclusion}

In this paper, we explore the head structure designing to improve the representation learning capabilities of DNNs for image classification. NOAH, a general-purposed classification head alternative built upon novel pairwise object category attentions at local to global scales, is presented. We show that NOAH can attain promising performance on different benchmarks using various DNN architectures. 

%
%
%



\bibliographystyle{IEEEtran}
\bibliography{example_paper}

\begin{thebibliography}{10}
\providecommand{\url}[1]{#1}
\csname url@samestyle\endcsname
\providecommand{\newblock}{\relax}
\providecommand{\bibinfo}[2]{#2}
\providecommand{\BIBentrySTDinterwordspacing}{\spaceskip=0pt\relax}
\providecommand{\BIBentryALTinterwordstretchfactor}{4}
\providecommand{\BIBentryALTinterwordspacing}{\spaceskip=\fontdimen2\font plus
\BIBentryALTinterwordstretchfactor\fontdimen3\font minus
  \fontdimen4\font\relax}
\providecommand{\BIBforeignlanguage}[2]{{%
\expandafter\ifx\csname l@#1\endcsname\relax
\typeout{** WARNING: IEEEtran.bst: No hyphenation pattern has been}%
\typeout{** loaded for the language `#1'. Using the pattern for}%
\typeout{** the default language instead.}%
\else
\language=\csname l@#1\endcsname
\fi
#2}}
\providecommand{\BIBdecl}{\relax}
\BIBdecl

\bibitem{sift}
D.~G. Lowe, ``Object recognition from local scale-invariant features,'' in
  \emph{ICCV}, 1999.

\bibitem{hog}
N.~Dalal and B.~Triggs, ``Histograms of oriented gradients for human
  detection,'' in \emph{CVPR}, 2005.

\bibitem{spatial_pyramid_matching}
S.~Lazebnik, C.~Schmid, and J.~Ponce, ``Beyond bags of features: Spatial
  pyramid matching for recognizing natural scene categories,'' in \emph{CVPR},
  2006.

\bibitem{alexnet}
A.~Krizhevsky, I.~Sutskever, and G.~E. Hinton, ``Imagenet classification with
  deep convolutional neural networks,'' in \emph{NIPS}, 2012.

\bibitem{vgg}
K.~Simonyan and A.~Zisserman, ``Very deep convolutional networks for
  large-scale image recognition,'' in \emph{ICLR}, 2015.

\bibitem{googlenet}
C.~Szegedy, W.~Liu, Y.~Jia, P.~Sermanet, S.~Reed, D.~Anguelov, D.~Erhan,
  V.~Vanhoucke, and A.~Rabinovich, ``Going deeper with convolutions,'' in
  \emph{CVPR}, 2015.

\bibitem{nin}
M.~Lin, Q.~Chen, and S.~Yan, ``Network in network,'' in \emph{ICLR}, 2014.

\bibitem{resnet}
K.~He, X.~Zhang, S.~Ren, and J.~Sun, ``Deep residual learning for image
  recognition,'' in \emph{CVPR}, 2016.

\bibitem{nas}
B.~Zoph and Q.~V. Le, ``Neural architecture search with reinforcement
  learning,'' in \emph{ICLR}, 2017.

\bibitem{regnet}
I.~Radosavovic, R.~P. Kosaraju, R.~Girshick, K.~He, and P.~Doll{\'a}r,
  ``Designing network design spaces,'' in \emph{CVPR}, 2020.

\bibitem{convnext}
Z.~Liu, H.~Mao, C.-Y. Wu, C.~Feichtenhofer, T.~Darrell, and S.~Xie, ``A convnet
  for the 2020s,'' in \emph{CVPR}, 2022.

\bibitem{vit}
A.~Dosovitskiy, L.~Beyer, A.~Kolesnikov, D.~Weissenborn, X.~Zhai,
  T.~Unterthiner, M.~Dehghani, M.~Minderer, G.~Heigold, S.~Gelly \emph{et~al.},
  ``An image is worth 16x16 words: Transformers for image recognition at
  scale,'' in \emph{ICLR}, 2021.

\bibitem{transformer}
A.~Vaswani, N.~Shazeer, N.~Parmar, J.~Uszkoreit, L.~Jones, A.~N. Gomez,
  {\L}.~Kaiser, and I.~Polosukhin, ``Attention is all you need,'' in
  \emph{NIPS}, 2017.

\bibitem{bert}
J.~Devlin, M.-W. Chang, K.~Lee, and K.~Toutanova, ``Bert: Pre-training of deep
  bidirectional transformers for language understanding,'' in \emph{NAACL},
  2019.

\bibitem{deit}
H.~Touvron, M.~Cord, M.~Douze, F.~Massa, A.~Sablayrolles, and H.~J{\'e}gou,
  ``Training data-efficient image transformers \& distillation through
  attention,'' in \emph{ICML}, 2021.

\bibitem{pvt}
W.~Wang, E.~Xie, X.~Li, D.-P. Fan, K.~Song, D.~Liang, T.~Lu, P.~Luo, and
  L.~Shao, ``Pyramid vision transformer: A versatile backbone for dense
  prediction without convolutions,'' in \emph{ICCV}, 2021.

\bibitem{swin}
Z.~Liu, Y.~Lin, Y.~Cao, H.~Hu, Y.~Wei, Z.~Zhang, S.~Lin, and B.~Guo, ``Swin
  transformer: Hierarchical vision transformer using shifted windows,'' in
  \emph{ICCV}, 2021.

\bibitem{shufflevit}
Z.~Huang, Y.~Ben, G.~Luo, P.~Cheng, G.~Yu, and B.~Fu, ``Shuffle transformer:
  Rethinking spatial shuffle for vision transformer,'' \emph{arXiv preprint
  arXiv:2106.03650}, 2021.

\bibitem{scalingvit}
X.~Zhai, A.~Kolesnikov, N.~Houlsby, and L.~Beyer, ``Scaling vision
  transformers,'' in \emph{CVPR}, 2022.

\bibitem{mlp}
L.~Melas-Kyriazi, ``Do you even need attention? a stack of feed-forward layers
  does surprisingly well on imagenet,'' \emph{arXiv preprint arXiv:2105.02723},
  2021.

\bibitem{mlpmixer}
I.~Tolstikhin, N.~Houlsby, A.~Kolesnikov, L.~Beyer, X.~Zhai, T.~Unterthiner,
  J.~Yung, D.~Keysers, J.~Uszkoreit, M.~Lucic \emph{et~al.}, ``Mlp-mixer: An
  all-mlp architecture for vision,'' in \emph{NeurIPS}, 2021.

\bibitem{resmlp}
H.~Touvron, P.~Bojanowski, M.~Caron, M.~Cord, A.~El-Nouby, E.~Grave,
  G.~Izacard, A.~Joulin, G.~Synnaeve, J.~Verbeek \emph{et~al.}, ``Resmlp:
  Feedforward networks for image classification with data-efficient training,''
  \emph{arXiv preprint arXiv:2105.03404}, 2021.

\bibitem{mobilenets}
A.~G. Howard, M.~Zhu, B.~Chen, D.~Kalenichenko, W.~Wang, T.~Weyand,
  M.~Andreetto, and H.~Adam, ``Mobilenets: Efficient convolutional neural
  networks for mobile vision applications,'' \emph{arXiv preprint
  arXiv:1704.04861}, 2017.

\bibitem{shufflenet}
X.~Zhang, X.~Zhou, M.~Lin, and J.~Sun, ``Shufflenet: An extremely efficient
  convolutional neural network for mobile devices,'' in \emph{CVPR}, 2018.

\bibitem{bipool}
T.-Y. Lin, A.~RoyChowdhury, and S.~Maji, ``Bilinear cnn models for fine-grained
  visual recognition,'' in \emph{ICCV}, 2015.

\bibitem{kernelpool}
Y.~Cui, F.~Zhou, J.~Wang, X.~Liu, Y.~Lin, and S.~Belongie, ``Kernel pooling for
  convolutional neural networks,'' in \emph{CVPR}, 2017.

\bibitem{saol}
I.~Kim, W.~Baek, and S.~Kim, ``Spatially attentive output layer for image
  classification,'' in \emph{CVPR}, 2020.

\bibitem{dataaug}
M.~A. Islam, M.~Kowal, S.~Jia, K.~G. Derpanis, and N.~D. Bruce, ``Global
  pooling, more than meets the eye: Position information is encoded
  channel-wise in cnns,'' in \emph{ICCV}, 2021.

\bibitem{resattention}
K.~Zhu and J.~Wu, ``Residual attention: A simple but effective method for
  multi-label recognition,'' in \emph{ICCV}, 2021.

\bibitem{query2label}
S.~Liu, L.~Zhang, X.~Yang, H.~Su, and J.~Zhu, ``Query2label: A simple
  transformer way to multi-label classification,'' \emph{arXiv preprint
  arXiv:2107.10834}, 2021.

\bibitem{sot}
J.~Xie, R.~Zeng, Q.~Wang, Z.~Zhou, and P.~Li, ``Sot: Delving deeper into
  classification head for transformer,'' \emph{arXiv preprint
  arXiv:2104.10935}, 2021.

\bibitem{market1501}
L.~Zheng, L.~Shen, L.~Tian, S.~Wang, J.~Wang, and Q.~Tian, ``Scalable person
  re-identification: A benchmark,'' in \emph{ICCV}, 2015.

\bibitem{inaturalist}
G.~Van~Horn, O.~Mac~Aodha, Y.~Song, Y.~Cui, C.~Sun, A.~Shepard, H.~Adam,
  P.~Perona, and S.~Belongie, ``The inaturalist species classification and
  detection dataset,'' in \emph{CVPR}, 2018.

\bibitem{coco}
T.-Y. Lin, M.~Maire, S.~Belongie, J.~Hays, P.~Perona, D.~Ramanan,
  P.~Doll{\'a}r, and C.~L. Zitnick, ``Microsoft coco: Common objects in
  context,'' in \emph{ECCV}, 2014.

\bibitem{densenet}
G.~Huang, Z.~Liu, L.~Van Der~Maaten, and K.~Q. Weinberger, ``Densely connected
  convolutional networks,'' in \emph{CVPR}, 2017.

\bibitem{mobilenetv2}
M.~Sandler, A.~Howard, M.~Zhu, A.~Zhmoginov, and L.-C. Chen, ``Mobilenetv2:
  Inverted residuals and linear bottlenecks,'' in \emph{CVPR}, 2018.

\bibitem{shufflenetv2}
N.~Ma, X.~Zhang, H.-T. Zheng, and J.~Sun, ``Shufflenet v2: Practical guidelines
  for efficient cnn architecture design,'' in \emph{ECCV}, 2018.

\bibitem{efficientnet}
M.~Tan and Q.~Le, ``Efficientnet: Rethinking model scaling for convolutional
  neural networks,'' in \emph{ICML}, 2019.

\bibitem{tnt}
K.~Han, A.~Xiao, E.~Wu, J.~Guo, C.~Xu, and Y.~Wang, ``Transformer in
  transformer,'' in \emph{NeurIPS}, 2021.

\bibitem{t2t}
L.~Yuan, Y.~Chen, T.~Wang, W.~Yu, Y.~Shi, Z.~Jiang, F.~E. Tay, J.~Feng, and
  S.~Yan, ``Tokens-to-token vit: Training vision transformers from scratch on
  imagenet,'' in \emph{ICCV}, 2021.

\bibitem{gmlp}
H.~Liu, Z.~Dai, D.~R. So, and Q.~V. Le, ``Pay attention to mlps,'' in
  \emph{NeurIPS}, 2021.

\bibitem{cyclemlp}
S.~Chen, E.~Xie, C.~Ge, R.~Chen, D.~Liang, and P.~Luo, ``Cyclemlp: A mlp-like
  architecture for dense prediction,'' in \emph{ICLR}, 2021.

\bibitem{s2mlp}
T.~Yu, X.~Li, Y.~Cai, M.~Sun, and P.~Li, ``S2-mlp: Spatial-shift mlp
  architecture for vision,'' in \emph{WACV}, 2022.

\bibitem{2vpmlp}
Q.~Hou, Z.~Jiang, L.~Yuan, M.-M. Cheng, S.~Yan, and J.~Feng, ``Vision
  permutator: A permutable mlp-like architecture for visual recognition,''
  \emph{TPAMI}, 2022.

\bibitem{stopool}
M.~D. Zeiler and R.~Fergus, ``Stochastic pooling for regularization of deep
  convolutional neural networks,'' in \emph{ICLR}, 2013.

\bibitem{combipool}
Y.~Gao, O.~Beijbom, N.~Zhang, and T.~Darrell, ``Compact bilinear pooling,'' in
  \emph{CVPR}, 2016.

\bibitem{gsop}
Z.~Gao, J.~Xie, Q.~Wang, and P.~Li, ``Global second-order pooling convolutional
  networks,'' in \emph{CVPR}, 2019.

\bibitem{gcp}
C.~Ionescu, O.~Vantzos, and C.~Sminchisescu, ``Matrix backpropagation for deep
  networks with structured layers,'' in \emph{ICCV}, 2015.

\bibitem{fastgcp}
P.~Li, J.~Xie, Q.~Wang, and Z.~Gao, ``Towards faster training of global
  covariance pooling networks by iterative matrix square root normalization,''
  in \emph{CVPR}, 2018.

\bibitem{detailpool}
F.~Saeedan, N.~Weber, M.~Goesele, and S.~Roth, ``Detail-preserving pooling in
  deep networks,'' in \emph{CVPR}, 2018.

\bibitem{liftpool}
J.~Zhao and C.~G.~M. Snoek, ``Liftpool: Bidirectional convnet pooling,'' in
  \emph{ICLR}, 2021.

\bibitem{blurpool}
R.~Zhang, ``Making convolutional networks shift-invariant again,'' in
  \emph{ICML}, 2019.

\bibitem{spectralpool}
O.~Rippel, J.~Snoek, and R.~P. Adams, ``Spectral representations for
  convolutional neural networks,'' in \emph{NIPS}, 2015.

\bibitem{s3pool}
S.~Zhai, H.~Wu, A.~Kumar, Y.~Cheng, Y.~Lu, Z.~Zhang, and R.~Feris, ``S3pool:
  Pooling with stochastic spatial sampling,'' in \emph{CVPR}, 2017.

\bibitem{lip}
Z.~Gao, L.~Wang, and G.~Wu, ``Lip: Local importance-based pooling,'' in
  \emph{ICCV}, 2019.

\bibitem{gcpjournal}
Q.~Wang, J.~Xie, W.~Zuo, L.~Zhang, and P.~Li, ``Deep cnns meet global
  covariance pooling: Better representation and generalization,'' \emph{TPAMI},
  2021.

\bibitem{gausspool}
T.~Kobayashi, ``Gaussian-based pooling for convolutional neural networks,'' in
  \emph{NeurIPS}, 2019.

\bibitem{crosslayerpool}
L.~Liu, C.~Shen, and A.~Van Den~Hengel, ``The treasure beneath convolutional
  layers: Cross-convolutional-layer pooling for image classification,'' in
  \emph{CVPR}, 2015.

\bibitem{mcar}
B.-B. Gao and H.-Y. Zhou, ``Learning to discover multi-class attentional
  regions for multi-label image recognition,'' \emph{TIP}, 2021.

\bibitem{harmoniousattention}
W.~Li, X.~Zhu, and S.~Gong, ``Harmonious attention network for person
  re-identification,'' in \emph{CVPR}, 2018.

\bibitem{netvlad}
R.~Arandjelovic, P.~Gronat, A.~Torii, T.~Pajdla, and J.~Sivic, ``Netvlad: Cnn
  architecture for weakly supervised place recognition,'' in \emph{CVPR}, 2016.

\bibitem{fishernet}
P.~Tang, X.~Wang, B.~Shi, X.~Bai, W.~Liu, and Z.~Tu, ``Deep fishernet for
  object classification,'' \emph{arXiv preprint arXiv: 1608.00182}, 2016.

\bibitem{g2denet}
Q.~Wang, P.~Li, and L.~Zhang, ``G2denet: Global gaussian distribution embedding
  network and its application to visual recognition,'' in \emph{CVPR}, 2017.

\bibitem{monet}
M.~Gou, F.~Xiong, O.~Camps, and M.~Sznaier, ``Monet: Moments embedding
  network,'' in \emph{CVPR}, 2018.

\bibitem{labelgraph}
X.~Zhu, J.~Liu, W.~Liu, J.~Ge, B.~Liu, and J.~Cao, ``Scene-aware label graph
  learning for multi-label image classification,'' in \emph{ICCV}, 2023.

\bibitem{spatialattention}
F.~Zhu, H.~Li, W.~Ouyang, N.~Yu, and X.~Wang, ``Learning spatial regularization
  with image-level supervisions for multi-label image classification,'' in
  \emph{CVPR}, 2017.

\bibitem{softaug}
Y.~Liu, S.~Yan, L.~Leal-Taix{\'e}, J.~Hays, and D.~Ramanan, ``Soft augmentation
  for image classification,'' in \emph{CVPR}, 2023.

\bibitem{residualattention}
F.~Wang, M.~Jiang, C.~Qian, S.~Yang, C.~Li, H.~Zhang, X.~Wang, and X.~Tang,
  ``Residual attention network for image classification,'' in \emph{CVPR},
  2017.

\bibitem{cbam}
S.~Woo, J.~Park, J.-Y. Lee, and I.~S. Kweon, ``Cbam: Convolutional block
  attention module,'' in \emph{ECCV}, 2018.

\bibitem{se}
J.~Hu, L.~Shen, and G.~Sun, ``Squeeze-and-excitation networks,'' in
  \emph{CVPR}, 2018.

\bibitem{imagenet}
O.~Russakovsky, J.~Deng, H.~Su, J.~Krause, S.~Satheesh, S.~Ma, Z.~Huang,
  A.~Karpathy, A.~Khosla, M.~Bernstein, A.~C. Berg, and F.-F. Li, ``Imagenet
  large scale visual recognition challenge,'' \emph{IJCV}, 2015.

\bibitem{mixed_gated}
C.-Y. Lee, P.~W. Gallagher, and Z.~Tu, ``Generalizing pooling functions in
  convolutional neural networks: Mixed, gated, and tree,'' in \emph{AISTATS},
  2016.

\bibitem{gfgp}
T.~Kobayashi, ``Global feature guided local pooling,'' in \emph{ICCV}, 2019.

\bibitem{3gnet}
Q.~Wang, P.~Li, Q.~Hu, P.~Zhu, and W.~Zuo, ``Deep global generalized gaussian
  networks,'' in \emph{CVPR}, 2019.

\bibitem{understandgcp}
Q.~Wang, L.~Zhang, B.~Wu, D.~Ren, P.~Li, W.~Zuo, and Q.~Hu, ``What deep cnns
  benefit from global covariance pooling: an optimization perspective,'' in
  \emph{CVPR}, 2020.

\bibitem{mlgcn}
Z.-M. Chen, X.-S. Wei, P.~Wang, and Y.~Guo, ``Multi-label image recognition
  with graph convolutional networks,'' in \emph{CVPR}, 2019.

\bibitem{agcn}
Q.~Li, X.~Peng, Y.~Qiao, and Q.~Peng, ``Learning label correlations for
  multi-label image recognition with graph networks,'' \emph{Pattern
  Recognition Letters}, 2020.

\bibitem{fgcn}
Y.~Wang, Y.~Xie, Y.~Liu, K.~Zhou, and X.~Li, ``Fast graph convolution network
  based multi-label image recognition via cross-modal fusion,'' in \emph{CIKM},
  2020.

\bibitem{cma}
R.~You, Z.~Guo, L.~Cui, X.~Long, Y.~Bao, and S.~Wen, ``Cross-modality attention
  with semantic graph embedding for multi-label classification,'' in
  \emph{AAAI}, 2020.

\bibitem{msrn}
X.~Qu, H.~Che, J.~Huang, L.~Xu, and X.~Zheng, ``Multi-layered semantic
  representation network for multi-label image classification,'' \emph{arXiv
  preprint arXiv:2106.11596}, 2021.

\bibitem{eca}
Q.~Wang, B.~Wu, P.~Zhu, P.~Li, W.~Zuo, and Q.~Hu, ``Eca-net: Efficient channel
  attention for deep convolutional neural networks,'' in \emph{CVPR}, 2020.

\bibitem{cgc}
X.~Lin, L.~Ma, W.~Liu, and S.-F. Chang, ``Context-gated convolution,'' in
  \emph{ECCV}, 2020.

\bibitem{weightnet}
N.~Ma, X.~Zhang, J.~Huang, and J.~Sun, ``Weightnet: Revisiting the design space
  of weight networks,'' in \emph{ECCV}, 2020.

\bibitem{we}
N.~Quader, M.~M.~I. Bhuiyan, J.~Lu, P.~Dai, and W.~Li, ``Weight excitation:
  Built-in attention mechanisms in convolutional neural networks,'' in
  \emph{ECCV}, 2020.

\bibitem{gradcam}
A.~Chattopadhay, A.~Sarkar, P.~Howlader, and V.~N. Balasubramanian,
  ``Grad-cam++: Generalized gradient-based visual explanations for deep
  convolutional networks,'' in \emph{WACV}, 2018.

\bibitem{adamw}
I.~Loshchilov and F.~Hutter, ``Decoupled weight decay regularization,'' in
  \emph{ICLR}, 2019.

\bibitem{labelsmoothing}
C.~Szegedy, V.~Vanhoucke, S.~Ioffe, S.~Jonathon, and W.~Zbigniew, ``Rethinking
  the inception architecture for computer vision,'' in \emph{CVPR}, 2016.

\bibitem{randaugment}
E.~D. Cubuk, B.~Zoph, J.~Shlens, and L.~Q. V, ``Randaugment: Practical
  automated data augmentation with a reduced search space,'' in \emph{NeurIPS},
  2020.

\bibitem{randerazing}
Z.~Zhong, L.~Zheng, G.~Kang, S.~Li, and Y.~Yang, ``Random erasing data
  augmentation,'' in \emph{AAAI}, 2020.

\bibitem{mixup}
H.~Zhang, C.~Moustapha, N.~D. Yann, and L.-P. David, ``mixup: Beyond empirical
  risk minimization,'' in \emph{ICLR}, 2018.

\bibitem{cutmix}
S.~Yun, , D.~Han, S.~J. Oh, S.~Chun, J.~Choe, and Y.~Yoo, ``Cutmix:
  Regularization strategy to train strong classifiers with localizable
  features,'' in \emph{ICCV}, 2019.

\bibitem{stochasticdepth}
G.~Huang, Y.~Sun, Z.~Liu, S.~Daniel, and W.~K. Q, ``Deep networks with
  stochastic depth,'' in \emph{ECCV}, 2016.

\bibitem{cait}
H.~Touvron, M.~Cord, A.~Sablayrolles, G.~Synnaeve, and H.~Jegou, ``Going deeper
  with image transformers,'' in \emph{ICCV}, 2021.

\bibitem{ema}
B.~T. Polyak and A.~B. Juditsky, ``Acceleration of stochastic approximation by
  averaging,'' \emph{SIAM Journal on Control and Optimization}, 1992.

\end{thebibliography}

%
%
%
%
%
%

\clearpage
\appendix

In this section, supplementary materials are provided. 



\subsection{Multi-class Image Classification on ImageNet Dataset}

ImageNet classification dataset consists of over 1.2 million images for training and 50,000 images for validation, including 1,000 image classes. To have a comprehensive evaluation conditioned on the extreme capability of our current computational resources, we apply NOAH to a variety of DNN architectures including 9 CNN backbones, 8 ViT backbones and 8 MLP backones, covering a relatively large range of model size. For CNNs, we select backbones from ResNet~\cite{resnet}, MobileNetV2~\cite{mobilenetv2}, and ShuffleNetV2~\cite{shufflenetv2} families. For ViTs, we select backbones from DeiT~\cite{deit} and PVT~\cite{pvt} families. For MLPs, we select backbones from Mixer~\cite{mlpmixer} and gMLP~\cite{gmlp} families. In the experiments, we construct our networks by replacing the existing head of each selected DNN architecture by a NOAH. Typically, we adopt the standard data augmentation to train and evaluate each network, unless otherwise stated. For training, we first resize the input images to $256\times256$, then randomly sample $224\times224$ image crops or their horizontal flips. We standardize the cropped images with mean and variance per channel. For evaluation, we use the center crops of the resized images, and report top-1 and top-5 recognition rates on the ImageNet validation set. For fair comparisons, we use the public PyTorch codes of these networks~\footnote{\url{https://github.com/pytorch/vision/tree/main/torchvision/models}}
~\footnote{\url{https://github.com/facebookresearch/deit}}~\footnote{\url{https://github.com/whai362/PVT}}~\footnote{\url{https://github.com/rwightman/pytorch-image-models}} with the exactly same settings to train all baseline models and our models from scratch.~\textit{Note that our trained baseline models are better than or at least on par with the reported ones}.

Specifically, the models of ResNet18, ResNet50, ResNet101, ResNet152, MobileNetV2 ($1.0\times$), MobileNetV2 ($0.75\times$), MobileNetV2 ($0.5\times$), MobileNetV2 ($0.35\times$), ShuffleNetV2  ($1.0\times$), DeiT-Tiny ($1.0\times$), DeiT-Tiny ($0.75\times$), DeiT-Tiny ($0.5\times$), PVT-Tiny ($1.0\times$), PVT-Tiny ($0.75\times$), PVT-Tiny ($0.5\times$), Mixer-Small ($1.0\times$), Mixer-Small ($0.75\times$), Mixer-Small ($0.5\times$), gMLP-Tiny ($1.0\times$), gMLP-Tiny ($0.75\times$) and gMLP-Tiny ($0.5\times$) are trained on servers with 8 NVIDIA Titan X GPUs. The other models of DeiT-Base, DeiT-Small, Mixer-Base and gMLP-Small, which require much larger memory cost, are trained on servers with 8 NVIDIA Tesla V100 GPUs. Detailed training setups for different DNN backbones are as follows.

\textbf{Training setup for ResNet models.} The initial learning rate is set to 0.1 and decayed by a factor of 10 every 30 epochs. All models are trained by the stochastic gradient descent (SGD) optimizer for 100 epochs, with a batch size of 256, a weight decay of 0.0001 and a momentum of 0.9.

\textbf{Training setup for MobileNet models.} The initial learning rate is set to 0.05 and scheduled to arrive at zero with a cosine decaying strategy. All models are trained by the SGD optimizer for 150 epochs, with a batch size of 256, a weight decay of 0.00004 and a momentum of 0.9.

\textbf{Training setup for ShuffleNetV2 models.} The initial learning rate is set to 0.5 and scheduled to arrive at zero linearly. All models are trained by the SGD optimizer for 240 epochs, with a batch size of 1024, a weight decay of 0.00004 and a momentum of 0.9.

\textbf{Training setup for DeiT and PVT models.} The initial learning rate is set to 0.0005 and scheduled to arrive at zero with a cosine decaying strategy. All models are trained by the AdamW optimizer~\cite{adamw} for 300 epochs, with a batch size of 1024, a weight decay of 0.05 and a momentum of 0.9. Following~\cite{deit} and~\cite{pvt}, we use Label Smoothing~\cite{labelsmoothing}, RandAugment~\cite{randaugment}, Random Erasing~\cite{randerazing}, Mixup~\cite{mixup} and CutMix~\cite{cutmix} during training.

\textbf{Training setup for Mixer and gMLP models.} The initial learning rate is set to 0.0007 and scheduled to arrive at zero with a cosine decaying strategy. All models are trained by the AdamW optimizer for 300 epochs, with a batch size of 1536, a weight decay of 0.067 and a momentum of 0.9. Following~\cite{mlpmixer} and~\cite{gmlp}, we use Label Smoothing, RandAugment, Random Erasing, Mixup and CutMix during training.

\subsection{Multi-label Image Classification on MS-COCO Dataset}

MS-COCO 2014 dataset~\cite{coco} consists of 82,081 images for training and 40,137 images for validation, including 80 object categories. We select ResNet101 (a commonly used benchmarking model for MS-COCO dataset) and ViT-Large~\cite{vit} architectures for experiments, and construct our models by replacing the existing head of each of them by a NOAH. We follow the experimental settings in~\cite{resattention,query2label}, and report mean average precision (mAP), per-category precision (CP), recall (CR), F1-score (CF1), overall precision (OP), recall (OR) and F1-score (OF1) for the overall statistics. 

\begin{figure*}[htbp]
\begin{minipage}[t]{0.45\linewidth}
\centering
\includegraphics[width=1.0\linewidth]{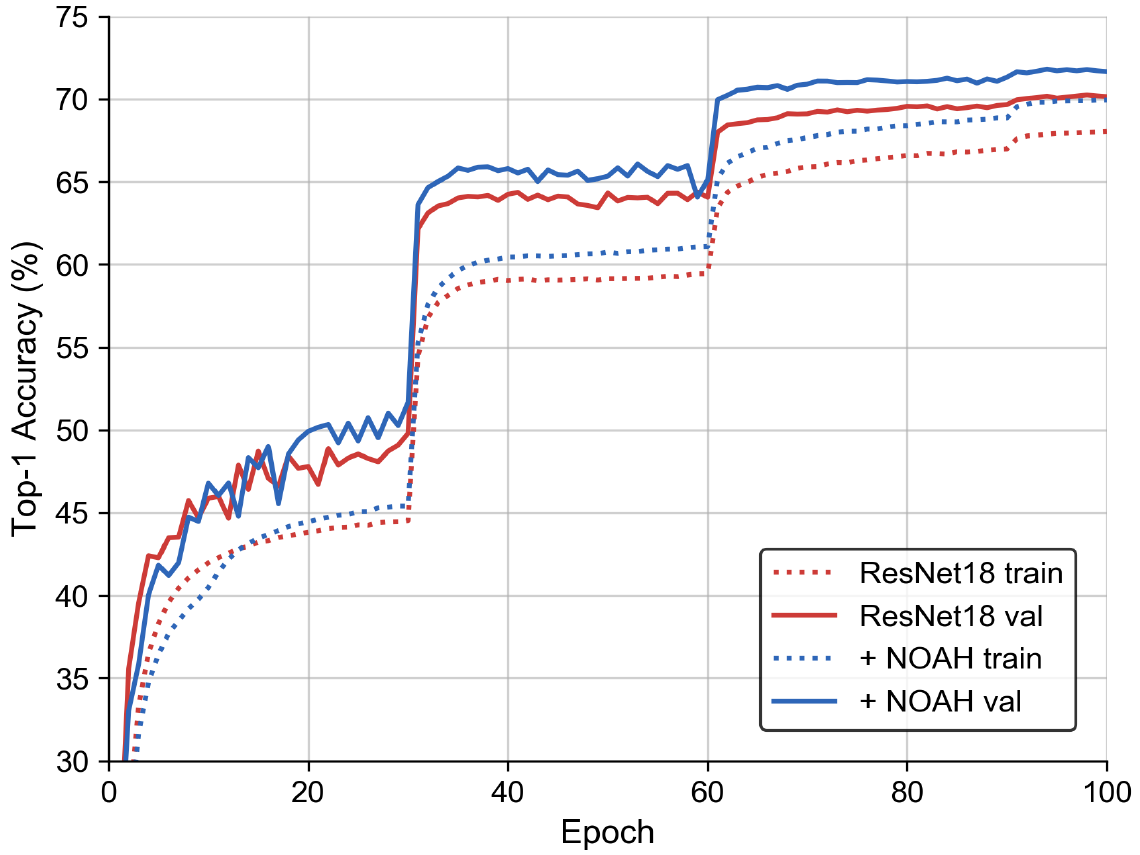}
\end{minipage}%
\hfill
\begin{minipage}[t]{0.45\linewidth}
\centering
\includegraphics[width=1.0\linewidth]{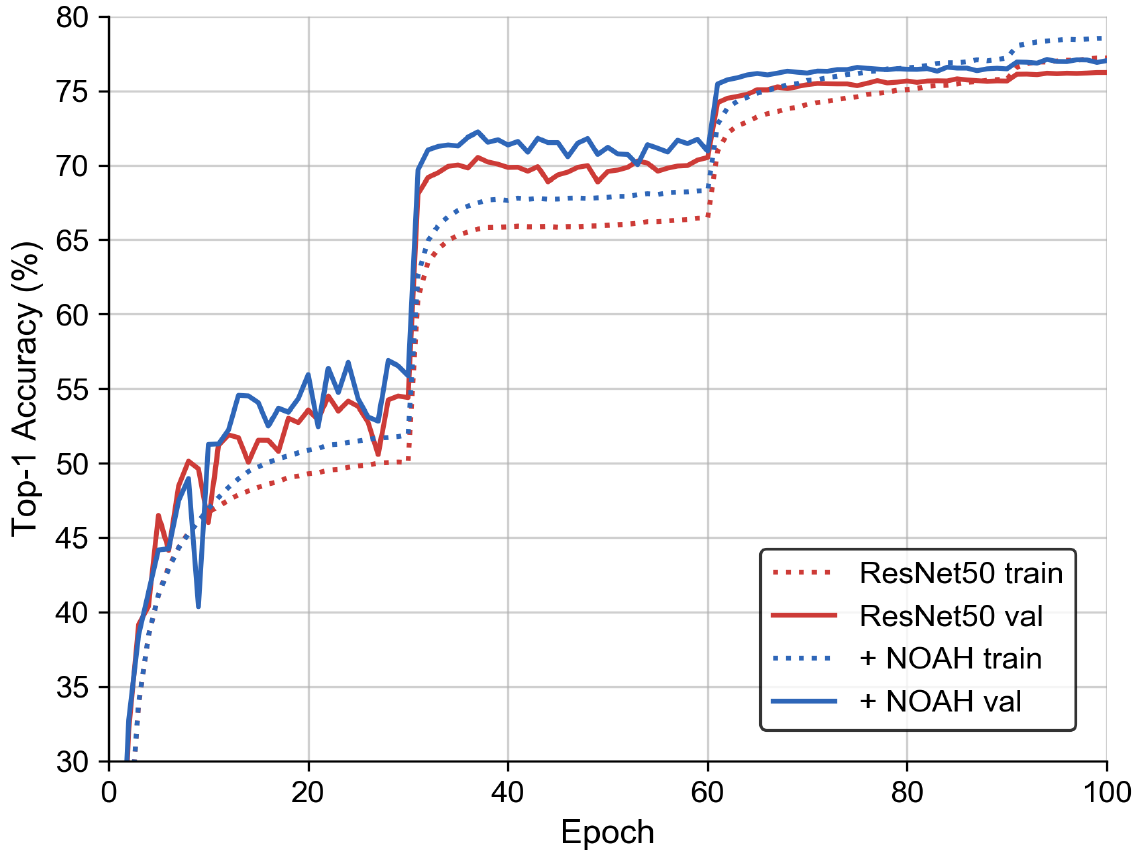}
\end{minipage}%

\begin{minipage}[t]{0.45\linewidth}
\centering
\includegraphics[width=1.0\linewidth]{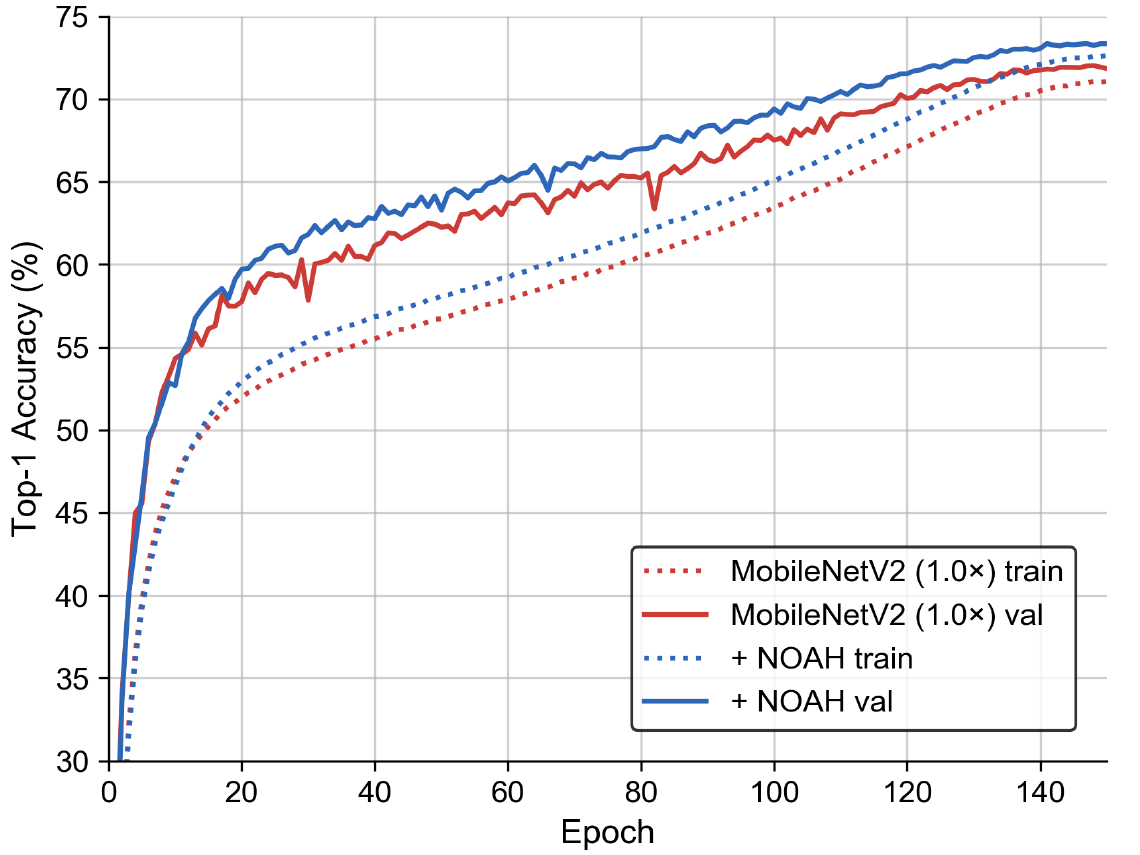}
\end{minipage}%
\hfill
\begin{minipage}[t]{0.45\linewidth}
\centering
\includegraphics[width=1.0\linewidth]{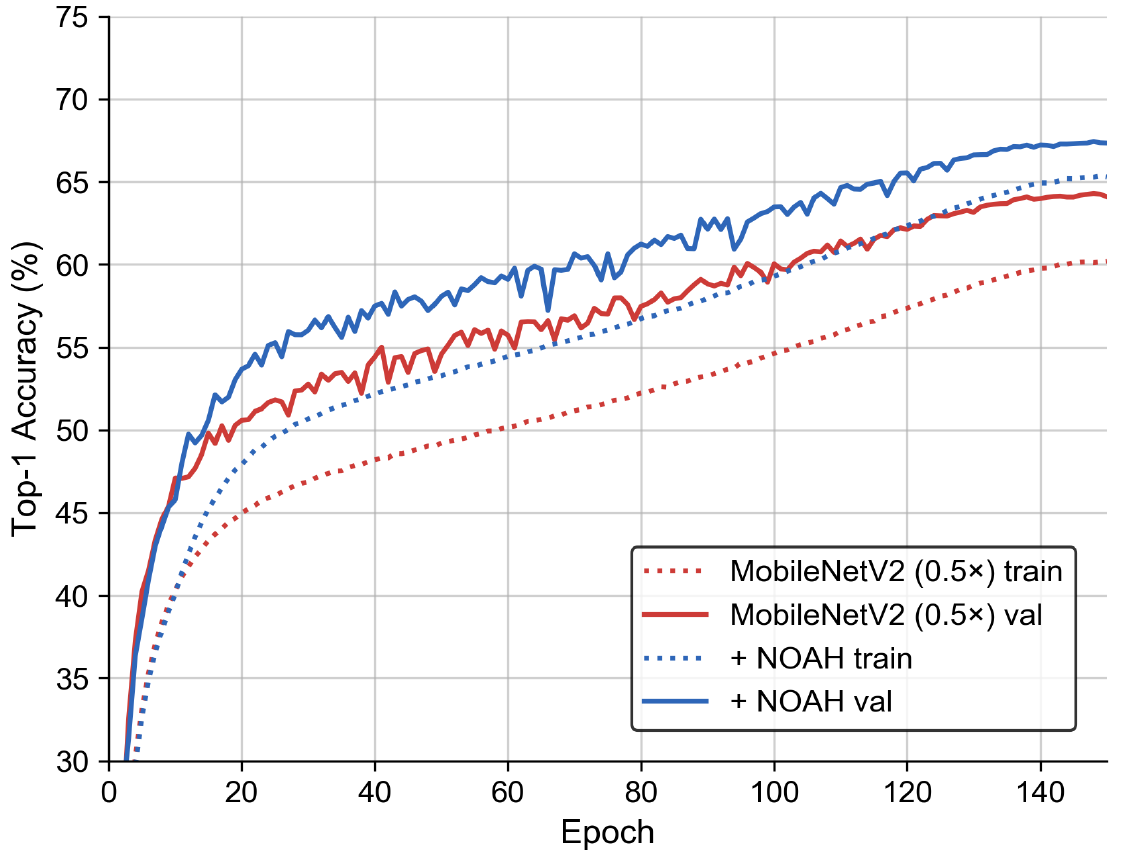}
\end{minipage}%

\centering
\vskip -0.in
	\caption{Curves of top-1 training accuracy (dashed line) and validation accuracy (solid line) of the ResNet18/ResNet50/MobileNetV2 ($1.0\times$)/MobileNetV2 ($0.5\times$) models trained on ImageNet dataset with the original head based on global feature encoding vs. NOAH. Comparatively, the ResNet18/ResNet50/MobileNetV2 ($1.0\times$)/MobileNetV2 ($0.5\times$) model with NOAH converges with the best validation accuracy, showing $1.56\%/1.02\%/1.33\%/3.14\%$ top-1 gain to the baseline while maintaining almost the same model size, respectively.}
	\label{fig:trainingcurves}
\vskip -0.in
\end{figure*}

\textbf{Fine-tuning setup for ResNet101 model.} The initial learning rate is set to 0.01 and decayed by a factor of 10 every 5 epochs. All models are trained by the stochastic gradient descent (SGD) optimizer for 30 epochs, with a batch size of 16, a weight decay of 0.0001 and a momentum of 0.9. For the first 2 epochs, we freeze the backbone weights (the models are all pre-trained on ImageNet dataset with a resolution of $224 \times 224$) and train the head only. For fine-tuning, we randomly flip each image for data augmentation. We resize all the images to $448\times448$ and standardize the cropped images with mean and variance per channel, for both fine-tuning and evaluation. For a fair comparison, we mainly compare NOAH with existing methods which report the results for the ResNet101 backbone with an image size of $448 \times 448$. All pre-trained models are fine-tuned with a single NVIDIA Titan X GPU.

\textbf{Fine-tuning setup for ViT-Large model.} The initial learning rate is set to 0.0001. All models are fine-tuned by the Adam optimizer for 80 epochs using one-cycle policy, with a batch size of 128, a weight decay of 0.01, and exponential moving average (EMA) with a decay of 0.9997. For fine-tuning, we use RandAugment for data augmentation. The models are all pre-trained on the ImageNet-22K classification dataset with a resolution of $384 \times 384$. We resize all the images to $448\times448$, standardize the cropped images with mean and variance per channel, for both fine-tuning and evaluation. All pre-trained models are fine-tuned on a server with 8 NVIDIA Titan X GPUs.

\subsection{Ablation: Fine-grained Image Classification on iNaturalist Dataset}

iNaturalist 2017 dataset~\cite{inaturalist} consists of 579,184 images for training and 95,986 images for validation, including 5,089 classes. Compared to ImageNet dataset, iNaturalist 2017 dataset has significantly more image classes and imbalanced class distribution. We select ResNet18 and ResNet50 as two backbones for experiments, and construct our models by replacing the existing head of each of them by a NOAH.

\textbf{Training setup for ResNet models.} The initial learning rate is set to 0.1 and decayed by a factor of 10 every 30 epochs. All models are trained from scratch by the stochastic gradient descent (SGD) optimizer for 100 epochs, with a batch size of 256, a weight decay of 0.0001 and a momentum of 0.9. For training, we first resize the input images to $256\times256$, then randomly sample $224\times224$ image crops or their horizontal flips. We standardize the cropped images with mean and variance per channel. For evaluation, we use the center crops of the resized images, and report top-1 recognition rate on the validation set. All models are trained on a server with 8 NVIDIA Titan X GPUs from scratch.

\subsection{Ablation: Person Re-identification on Market-1501 Dataset}

Market-1501~\cite{market1501} consists of 750 and 751 identities for training and testing, respectively. We adopt ResNet50 as the backbone, following the common settings on Market-1501. Specifically, an extra FC layer is appended after the GAP layer of ResNet50 first, then the output 512-D feature vector is used for person matching. We consider two training regimes: the standard from-scratch training and the fine-tuning. All models are trained with a single NVIDIA Titan X GPU.

\textbf{From-scratch training.} The initial learning rate is set to 0.065 and decayed by a factor of 10 at epoch 150, 225 and 300. All models are trained by the SGD optimizer for 350 epochs, with a batch size of 64, a weight decay of 0.0001 and a momentum of 0.9.

\textbf{Fine-tuning.} The initial learning rate is set to 0.0003 and decayed by a factor of 10 every 60 epochs. All models pre-trained on ImageNet are fine-tuned by the Amsgrad optimizer (in PyTorch) for 150 epochs, with a batch size of 64, a weight decay of 0.0001 and a momentum of 0.9.

\subsection{Ablation: Effect of the Aggressive Training Regime}

Recent work ConvNeXt~\cite{convnext} shows that, more accurate models can be attained when properly using much more aggressive augmentations, compared to the standard training regime. We also conduct ablative experiments to explore the generalization ability of NOAH in this aggressive from-scratch model training regime. Specifically, based on the public code~\footnote{\url{https://github.com/facebookresearch/ConvNeXt}} with the default settings, we compare the training of ResNet50 on ImageNet with vs. without using NOAH. We use the AdamW optimizer to train each model for 300 epochs with a learning rate of 0.004, a batch size of 4096 and a weight decay of 0.05. There is a 20-epoch linear warmup and a cosine learning rate decaying schedule afterward. For data augmentations, popular schemes including Mixup, CutMix, RandAugment and Random Erasing are used. Besides, Stochastic Depth~\cite{stochasticdepth}, Label Smoothing, LayerScale~\cite{cait} and Exponential Moving Average~\cite{ema} are also adopted to regularize the training process.

\subsection{Training Stability of NOAH}

Figure~\ref{fig:trainingcurves} shows the training and validation accuracy curves of the ResNet18/ResNet50/MobileNetV2 ($1.0\times$)/MobileNetV2 ($0.5\times$) models trained on ImageNet dataset with the original head based on global feature encoding vs. NOAH. We can observe that the ResNet18/ResNet50/MobileNetV2 ($1.0\times$)/MobileNetV2 ($0.5\times$) model with NOAH shows relatively high top-1 gains throughout the training process compared to the baseline, respectively.

\end{document}